\documentclass[final,authoryear,nopreprintline]{elsarticle}

\usepackage{lineno,hyperref}
\modulolinenumbers[5]
\usepackage{array,multirow,graphicx}
\usepackage{float}
\usepackage[dvipsnames]{xcolor}  
\usepackage{soul}  

\makeatletter
\newcommand\footnoteref[1]{\protected@xdef\@thefnmark{\ref{#1}}\@footnotemark}
\makeatother

\usepackage{caption}
\usepackage{subcaption}

\journal{Expert Systems with Applications}

\usepackage{appendix}
\usepackage{amsmath}
\usepackage{graphicx}
\usepackage{array}
\usepackage{longtable}
\usepackage{natbib}
\usepackage[utf8]{inputenc}
\usepackage[greek,english]{babel}
\usepackage{alphabeta}
\usepackage{amssymb}
\usepackage{latexsym}
\usepackage{footnote}
\makesavenoteenv{tabular}
\makesavenoteenv{table}
\newcommand*\patchAmsMathEnvironmentForLineno[1]{%
\expandafter\let\csname old#1\expandafter\endcsname\csname #1\endcsname
\expandafter\let\csname oldend#1\expandafter\endcsname\csname end#1\endcsname
\renewenvironment{#1}%
{\linenomath\csname old#1\endcsname}%
{\csname oldend#1\endcsname\endlinenomath}}%
\newcommand*\patchBothAmsMathEnvironmentsForLineno[1]{%
\patchAmsMathEnvironmentForLineno{#1}%
\patchAmsMathEnvironmentForLineno{#1*}}%
\AtBeginDocument{%
\patchBothAmsMathEnvironmentsForLineno{equation}%
\patchBothAmsMathEnvironmentsForLineno{align}%
\patchBothAmsMathEnvironmentsForLineno{flalign}%
\patchBothAmsMathEnvironmentsForLineno{alignat}%
\patchBothAmsMathEnvironmentsForLineno{gather}%
\patchBothAmsMathEnvironmentsForLineno{multline}%
}

\usepackage{cleveref}
\crefformat{section}{\S#2#1#3}
\crefformat{subsection}{\S#2#1#3}
\crefformat{subsubsection}{\S#2#1#3}
\crefrangeformat{section}{\S\S#3#1#4 to~#5#2#6}
\crefmultiformat{section}{\S\S#2#1#3}{ and~#2#1#3}{, #2#1#3}{ and~#2#1#3}

\begin{document}

\begin{frontmatter}

\title{A Neural Entity Coreference Resolution review}

\author[authcsd]{Nikolaos Stylianou\corref{mycorrespondingauthor}}
\cortext[mycorrespondingauthor]{Corresponding author at: School of Informatics, Aristotle University of Thessaloniki, Thessaloniki, 54124, Greece.}
\ead{nstylia@csd.auth.gr}

\author[authcsd]{Ioannis Vlahavas}
\ead{vlahavas@csd.auth.gr}

\address[authcsd]{School of Informatics, Aristotle University of Thessaloniki, Greece}

\begin{abstract}
Entity Coreference Resolution is the task of resolving all mentions in a document that refer to the same real world entity and is considered as one of the most difficult tasks in natural language understanding. It is of great importance for downstream natural language processing tasks such as entity linking, machine translation, summarization, chatbots, etc. This work aims to give a detailed review of current progress on solving Coreference Resolution using neural-based approaches. It also provides a detailed appraisal of the datasets and evaluation metrics in the field, as well as the subtask of Pronoun Resolution that has seen various improvements in the recent years. We highlight the advantages and disadvantages of the approaches, the challenges of the task, the lack of agreed-upon standards in the task and propose a way to further expand the boundaries of the field.

\end{abstract}


\end{frontmatter}


\section{Introduction}
\label{intro}
In everyday life we use language in many shapes and forms so as to express our thoughts and communicate. In order to successfully transfer our train of thought to another person, we have to be coherent. These coherent structures, which can be represented by a set of sentences when in writing, are commonly referred to as Discourse. 

When the events described are not presented in a linear way, achieving coherence can be challenging. While a structure can be coherent even if it does not follow the order of the events that took place, in order to ensure coherence in such structures, we have to make sure that they are cohesive, i.e. the way they are linked is meaningful.

There are many different forms of discourse. Texts, much like this one, are identified as monologues, while conversations are by definition dialogues of two or more participants. With the advancements in technology we also have Human-Computer Interaction (HCI), usually in the form of a human interacting with a bot (e.g. Siri on iOS devices). 

A very common linguistic phenomenon that we identify in all forms of language communication is when two expressions are used to refer to the same entity. We call this phenomenon an \textit{anaphora}, and the terms used to express this phenomenon as \textit{anaphoric}. The term anaphora originates from the Greek word ``αναφορά'', the etymology of which derives from the two words ``ανα'' and ``φορά''. The first term is a preposition indicating a temporal event, something that happened in the past, while the second term means ``to carry'', together forming a word which indicates that one is carrying something from before; a past term. 

The anaphoric terms are also defined as mentions or referring expressions, and the entity they refer to as referent \citep{Deemter-100}. The entity that the referring expression points to is defined as an antecedent. When two or more mentions refer to the same entity, the mentions corefer. Similarly, if an entity has only a single mention in a text, it is defined as a singleton. 

In Computational Linguistics, Coreference Resolution and Anaphora Resolution are the tasks that deal with the resolution of referring expressions and their antecedents \citep{poesio2016anaphora}. Coreference Resolution is the task of identifying different terms that refer to the same real world entity (corefer). Anaphora Resolution aims towards identifying the antecedent of an anaphoric pronoun or noun phrase in a text. 

To fully understand this field and analyze deep learning approaches towards solving these tasks we need to explore the different types of referring expressions that exist as well as the constrains that identify when two referring expressions can be linked. For that, it is crucial to identify what is considered as Coreference Resolution and what Anaphora Resolution in computational linguistics.

This review aims towards presenting in detail the field of Entity Coreference Resolution from a Deep Learning perspective as well as identifying the difficulties and innovations that past literature has. In the field of Anaphora Resolution, \cite{Mitkov1999ANAPHORAR} provides a detailed analysis of the state-of-the-art methods to that date, and an in depth explanation of the types of anaphora. A review on Coreference Resolution by \cite{Ng:2010:SNP:1858681.1858823} captures the first fifteen years of research on the field, explaining the variety of approaches that were implemented to date. The most recent survey by \cite{DBLP:journals/corr/abs-1805-11824} aims to summarize the advancements in both the fields of Coreference Resolution and Anaphora Resolution. {Although it touches on deep learning approaches briefly, it does not provide an in depth analysis of the methodologies and offers no semantic categorization of the approaches applied in the field. Moreover, recent developments in Natural Language Processing, such as the introduction of Transformers }\citep{vaswani2017attention}{, have signalled a new era of models with novel perspectives and architectures that have not been part of previous surveys. What is more, previous research only briefly reviews the resources used in Coreference Resolution, without analyzing their effects to the task.}

{We successfully bridge this gap with our work by providing a detailed review of neural Entity Coreference Resolution and its methodologies. To that end, we present the challenges of the task as a whole, identify the limitations of each methodology and provide insight on the developed approaches. We further provide an overview of the field, highlight the boundaries between Coreference Resolution and Anaphora Resolution, identify all resources used in the task and their characteristics and showcase the effect of the resources and their limitations to the task's progression. We also provide a brief outline of the different metrics used in the task and their shortcomings. Furthermore, we provide an extended synopsis of Pronoun Resolution, a subtask of Coreference Resolution that has been the subject of recent research due to it's importance to downstream tasks such as Machine Translation and Entity Linking, while emphasising on the work that been carried out towards gender bias. We believe that this detailed review will provide the necessary foundation to solve the lack of agreed-upon standards for Coreference Resolution by exhaustively examining all aspects of the task.}  

{For the purpose of this study, we reviewed 152 papers, published at peer-reviewed journals and conferences including available preprints and books. We have exhaustively identified all relative research via a combination of the PRISMA approach }\citep{moher2009preferred}{ according to which we applied keyword searching to individual publishers, ScienceDirect\footnote{\href{https://www.sciencedirect.com/}{https://www.sciencedirect.com/}}, Google Scholar\footnote{\href{https://scholar.google.com/}{https://scholar.google.com/}} and Arxiv\footnote{\href{https://arxiv.org/}{https://arxiv.org/}} and by incrementally discovering publications that have cited the previously identified works. We then accessed the collected publications in terms of quality and eligibility.In doing so we ensured that all works are directly related to Coreference Resolution, of high quality and have been peer reviewed. The preprints included are either accepted publications at conferences yet to be released in the proceedings or heavily cited publications that have been reviewed by the scientific community. As a result, the included publications consist of 22 journal articles, 105 conference papers, 6 books and 19 preprints.} 

In the following sections we describe in detail the anaphoric types that can be found in the English language (Section \ref{sec:Anaphoric-types}) as well as the constrains that apply on each mention in order to be matched with an antecedent (Section \ref{sec:constrains}). An overview of the differences between Coreference Resolution and Anaphora Resolution follows (Section \ref{sec:differences}). We then provide an apposition of the available datasets (Section \ref{sec:datasets}) and an overview of the evaluation metrics (Section \ref{sec:metrics}) for the aforementioned tasks, focusing on the advantages and disadvantages of each one. Section \ref{sec:CRhistory} provides a brief history of the non-neural approaches on Coreference Resolution, followed by Sections \ref{sec:ECR} \& \ref{sec:PR} which provide a detailed insight in the advancements in the tasks of Coreference Resolution and Pronoun Resolution, enabled by neural networks. We separate the approaches based on the methodologies and model types that were developed for Entity Coreference Resolution, while for Pronoun Resolution we differentiate between general and gender bias oriented works. {In Sections }\ref{sec:results}{, }\ref{sec:discussion}{ we present the results of the described approaches and offer remarks about their performance. Ultimately, Sections }\ref{sec:applications}{, }\ref{sec:challenges}{, }\ref{sec:conclusions}{, discuss the applications of Coreference Resolution in other tasks and the challenges that current systems face in an academic and practical setting, set the tone for future works and provide the final remarks of this study. }

\section{Anaphoric Types}
\label{sec:Anaphoric-types}
A variety of anaphoric types have been described in \citep{Hirst1981AnaphoraIN} and \citep{Lappin:1994:APA:203987.203989}. In the latter, those anaphoric types have been further expanded to include more distinct cases \citep{Mitkov1999ANAPHORAR,Ng:2010:SNP:1858681.1858823,Jurafsky:2009:SLP:1214993}. Different types of anaphora are resolved in the process of Coreference Resolution and different on Anaphora Resolution, while some are overlapping between the two.

Therefore the difference in the tasks of Coreference Resolution and Anaphora Resolution can be described by the anaphoric types that each one handles and their approach to resolve them. Many deep learning approaches handle specific anaphoric types, or excel in a certain anaphoric set, while struggle to resolve others. This can also be attributed to datasets that have not been well designed or do not hold a sufficient amount of anaphoric examples of some categories, as described in section \ref{sec:datasets},leading to the need for more targeted approaches. 

In this section, we list the different anaphoric types in the English language, give a brief explanation of their unique characteristics and differentiate them in terms of their information types.
We first start with the anaphoric types that point back to a certain antecedent. 

\paragraph{Zero Anaphora}
This type of anaphora makes use of a gap in a phrase or clause to point back to the antecedent. In most cases, the meaning of such an anaphora can only be understood by the extralinguistic context.
\paragraph{One Anaphora}
The type of anaphora is realized by the use of the word ``one'' in a noun phrase.
\paragraph{Pronominal Anaphora}
Considered to be the most widespread type, it is realized by the use of anaphoric pronouns and can be divided in three types, definite, indefinite and adjectival. The types of pronominal anaphora identify the type of the antecedent. In the case of definite, it refers to a single entity. Indefinite refers to an entity that is not well defined (e.g. ``group of people''). Adjectival refers to an entity that is described with the use of an adjective (e.g. ``good person''). The common factor in all of the pronominal anaphoras is that they refer to unique entities in the world, using a different approach of identifying it. 

\paragraph{Demonstratives} 
This is the type of anaphora that is used, as the name suggests, to demonstrate a certain entity in a comparative. Pronouns that express such behaviour are ``this'' and ``that''. These pronouns are further divided as proximal (for ``this'') and distal (for ``that''), as in their use they tend to showcase a certain distance; either literal or in time.

\paragraph{Presuppositions}
Refers to the ambiguous anaphoric pronouns such as someone, something, anyone, anything, etc.

\paragraph{Discontinuous Sets (Split Anaphora)}
This type refers to the anaphora in which the pronoun points back to more than one antecedent. 
\\

The anaphoric types can be further divided in categories based on the information structure they adhere to \citep{HAVILAND1974512, Prince_Taxonomy81, Nissim2004}.

\paragraph{Inferrable Anaphora (Bridging Anaphora)}
Also called \textit{bridging anaphora}, is a very specific anaphoric type that points back to another anaphoric phrase or clause, which in turn points to an entity mentioned further back in the document. 
\paragraph{Generics}
It refers to the case where a certain anaphoric term and its antecedent are not referring to the same real world entity. This behaviour makes such anaphoras to be more suitable members for the field of Coreference Resolution than Anaphora Resolution. 

\paragraph{Non referential terms}
It is very important in all the approaches of either Anaphora Resolution or Coreference Resolution to identify the anaphoric terms that are not pointing back to any antecedent. The pronoun ``it'' being the most common referring term that exhibits such behaviour. When found in syntactic phenomenons such as clefts and extrapositions it serves to suggest a certain behaviour while in other cases it is just pleonastic.  \\

A special anaphoric case exists, called \textit{cataphora}, in which the anaphoric term proceed the antecedent. All cataphora cases belong to one of the aforementioned anaphoric types with the antecedent appearing before the anaphoric noun phrase, hence pointing forward. 

\section{Anaphoric Constraints}
\label{sec:constrains}
In order to identify the right antecedent for each anaphoric noun phrase in a machine learning approach, certain syntactic and semantic features, that are meant to limit the possible referents, have to be considered. While these features have been implemented in different variations, they all serve the purpose of enforcing the linguistic properties that need to be satisfied before a link between a referent and a referring expression can be made.

\paragraph{Gender agreement}
Referents must agree with the gender of the referring expression to be considered as candidates.

\paragraph{Person agreement}
This constraint refers to the English form of person, which is split in three categories: first, second and third person. The third person is then identified as male, female or nonpersonal (``it'') gender. Referents and referring expressions must also match in this aspect to be linked. 

\paragraph{Number agreement}
Referents and referring expressing must agree in numbers, meaning that the expressions must be distinguished in singular and plural expressions. 

\paragraph{Binding theory}
Refers to the syntactic relationships that exist in English between the referential expressions and the possible antecedents when they appear in the same sentence, as identified by \cite{Chomsky1981}. One way to interpret this binding theory is to identify that reflexive pronouns (i.e. ``himself'', ``herself'', ``themselves'') co-refer with the immediate clause that contains them, whereas the opposite happens for non reflexive pronouns. 

\paragraph{Selectional Restrictions} This constraint refers to the elimination of a certain antecedent based on the properties a verb places on it. 

\paragraph{Recency} The referent of a pronoun is more likely to be introduced in statements that are closer to it. Therefore, we consider those clauses closer to the anaphoric pronouns more important. 
\\

Apart from the English syntactic constraints, a lot of earlier research has identified some not as strict constraints that set a priority of particular antecedents over others \citep{Jurafsky:2009:SLP:1214993}.

\paragraph{Discourse structure}
Limitations to the referent of an entity can be applied due to structural characteristics. 

The grammatical role of entities in subject position are more important than those in object position and as a result this behaviour translates to the mentions in subsequent positions. This is based on the salience hierarchy theory by \cite{Hajicova:1982:RHA:991813.991830}. 

Entities that have been repeatedly mentioned in the document, or have been focused on in prior discourse, are more likely to continue being treated the same way and therefore have a higher priority of being the referent. 

However, entities that appear in a parallelism phenomenon are more likely to ignore the grammatical role hierarchy constrains that were described above. 

The implicit causality of the verbs, as studied by \cite{CARAMAZZA1977601}, changes the properties of what is considered to be a subject and what an object. This, called verb semantics, differs from Selectional Restrictions as both possibilities are viable even after the restrictions a verb can place on it's arguments. Consider the following example: 
\\

George borrowed his car and his phone to Nick. He drove it to work.
\\

In this example, the selectional restrictions applies limit onto the ``it'', resulting in the pronoun only being able to refer to the car, instead of the phone as well. Similarly, verb semantics limit the antecedent selections from ``He'' to Nick instead of George based on the semantic emphasis the verb ``borrow'' applies.

\paragraph{World Knowledge}
This constraint is especially true in the task of Coreference Resolution, where terms refer to real world entities. This being the hardest constraint to incorporate into systems, since it is beyond syntactical and semantical constraints previously mentioned. As an example, if we consider the referring term ``the President'', in a current article about the United States, we need to have prior knowledge that Donald Trump is the current president in order to make the necessary reffering term - referent clustering. If the article was, however, a decade old, the term would be referring to Barack Obama, the U.S. President at the time. 

\section{Differences of Anaphora Resolution and Coreference Resolution}
\label{sec:differences}
Anaphora Resolution can be viewed as a subtask of Coreference Resolution, especially in the case of pronominal Anaphora Resolution which focuses on finding the antecedent of a pronoun to the nominal entity it points back to. However, there are cases where anaphora exists but not coreference \citep{Kibble2000CoreferenceAW}.
Futhermore, Anaphora Resolution is targeted towards identifying intra-linguistically determinable relations, while Coreference Resolution requires extra-linguistic information \citep{Kempson1977-KEMST}\footnote{Extra-linguistic information refers to factors that could affect the meaning based on parameters outside the text at hand, such as the historical period, world knowledge or a certain event. In comparison, intra-linguistic information can be found within the boundaries of the document.}. As a result, cases that require world knowledge are only resolved by Coreference Resolution, such as singletons (cases where a referring expression has no other mention in the document). What is more, Anaphora Resolution cannot solve phenomena like zero anaphora and discontinuous sets, while Coreference Resolution does not resolve Generics.

\section{Standard Datasets for Coreference Resolution}
\label{sec:datasets} 
A plethora of datasets have been developed to be used for the task of Coreference Resolution through the years.
In this survey we shall explore the CoNLL 2012 dataset \citep{Pradhan:2012:CST:2391181.2391183} which is predominantly used as the benchmark dataset in all state-of-the-art implementations for Entity Coreference Resolution as well as the GAP dataset \citep{GapDataset} that was developed towards Gender Ambiguous Pronoun Coreference Resolution and briefly compare their key factors with the other datasets, created to counter issues within the CoNLL dataset.

Historically, the first datasets that were created are the MUC 6 \citep{muc6:grishman1996message} and the MUC 7 \citep{muc7:chinchor1998overview}, developed for the 6$^{th}$ and 7$^{th}$ Message Understanding Conference respectively, being the ones that defined the task of Coreference Resolution. The datasets focus more on identity Coreference Resolution, while they do not contain annotations for binding theory coreference for example. 
The MUC datasets are relatively small in size, available only for the English language, and are homogeneous (i.e. all the documents are domain specific, in this case news articles). In the following decade, the ACE datasets \citep{ACE:L04-1011} were developed as part of the Automatic Content Extraction program to deal with the shortcomings of the MUC datasets. However, due to the program running for many years, four different versions of the ACE dataset have been developed (ACE02, ACE03, ACE04 and ACE05), initially including only news articles like the MUC datasets and then extending to other domains such as telephonic speech and broadcast conversations, as well as other languages (Chinese and Arabic). ACE datasets are also restricted in the semantic types that they annotate for this task. As a result, system evaluations and comparisons using the ACE datasets have been proved to be difficult. To make things worse, the ACE datasets do not have a specified train-test splits as the program organizers have not released the official test splits, making results on the ACE datasets even less trustworthy.

The CoNLL 2011 \citep{CoNLL11:W11-1901} and 2012 \citep{Pradhan:2012:CST:2391181.2391183} shared tasks included tasks on English Coreference Resolution and Multilingual Coreference Resolution respectively. The CoNLL 2011 dataset is based on OntoNotes 2.0 \citep{weischedel2007ontonotes}, while the CoNLL 2012 dataset is based on the OntoNotes 5.0 corpus \citep{weischedel2013ontonotes} and is aimed towards unrestricted Coreference Resolution. This resulted in the dataset having significantly more documents in the training and testing splits and being more robust than it's predecessors. Principally, this resulted in it arising as the benchmark dataset for all state-of-the-art work to this date. Nonetheless, research showcases that a high amount of overlap of previously seen mentions exists between the described splits \citep{moosavi-strube-2017-lexical} which leads to overfitting problems. 

In the following years, datasets were created to target task-specific resolution of Coreference and Anaphora. The ECB+ \citep{ecbplus:cybulska2014guidelines} dataset was introduced to handle topic-based event Coreference Resolution, while the ParCor \citep{ParCorDataset} dataset was aimed towards parallel pronoun Coreference Resolution to be later used for Machine Translation. Also, the CIC \citep{CicDataset:W16-3612} dataset had annotated multi-party conversations and was aimed to improve Coreference Resolution on Chatbots. ParCorFull \citep{lapshinova-koltunski-etal-2018-parcorfull}, an extention of the ParCor dataset also became available, with coreference annotations past pronouns.

In the more recent years, datasets have been created to tackle specific areas of Coreference Resolution as well, since the CoNLL 2012 dataset was lacking in either variety, or was missing a coreferring type completely. WikiCoref \citep{ghaddar2016wikicoref} has been proposed, which provides an unrestricted Coreference Resolution corpus, GUM \citep{GumDataset} was designed to handle domain adaptation, KnowRef \citep{emami-etal-2019-knowref} was developed to test world knowledge, while PreCo \citep{PreCoDataset} strives to improve error handling by providing separated analysis of mention detection and mention clustering. {LitBank }\citep{Bamman2020AnAD}{, which contains four times longer documents than CoNLL 2012, handles singletons and contains hard coreference annotations, is designed to server as a cross-domain and long-distance coreference performance benchmark.}

Gender bias has been a significant issue in the latest years, apparent in all fields of Computational Linguistics, spanning from text representation approaches to Coreference Resolution. In Coreference Resolution the bias appeared favouring the Masculine over Feminine predictions. As the bias originated in the datasets, the need to have unbiased training data resulted in the introduction of new sources. WinoBias \citep{zhao-etal-2018-gender} was introduced as the the first attempt on removing the bias of the CoNLL 2012 dataset. To enable a more targeted approach, Mind the GAP \citep{GapDataset} was published which aimed towards Gender Ambiguous Pronoun identification. As a result, it was recently used by Kaggle's Gender Pronoun Resolution competition\footnote{https://www.kaggle.com/c/gendered-pronoun-resolution}, and it is considered to be the benchmark for the task. In comparison to the CoNLL 2012 dataset, since it targets only gender specific pronoun resolution, the annotations scheme is very different - effectively shifting the clustering problem to a binary classification prediction. During the course of the competition it was also discovered that a small amount of the entries were miss-labelled\footnote{https://www.kaggle.com/c/gendered-pronoun-resolution/discussion/81331}, which are, to our best of knowledge, not fixed as of yet in the available version.  

All of the aforementioned corpora are composed of Prose type documents. However, there are differences that allow us to categorise them further. According to \cite{biber2019register}, we also distinguish texts based on their variety, register and dialects. As as result we introduce three types of categories: Media, Texts, and Conversational, based on the text styles (complete texts and text excerpts), and their difference in variety and register (academic papers versus face-to-face conversations). 

Media and Texts refer to complete texts, with Media consisting of documents from news articles or transcripts, blog posts and guide types, and Texts consisting of documents from academic writing, biographies, fiction, books and Wikipedia contents. Conversational consists of documents with exclusively multiple participants, such as forum discussions, talk shows, dialogues and telephone speech logs. What is more, Media and Texts are also different in terms of register, with Texts presuming a higher educational threshold.

\begin{table}[ht]
\centering
\resizebox{\textwidth}{!}{
\begin{tabular}{c|cc}
    Dataset  &  Categories & Annotation format \\
    \hline
    CoNLL \citep{Pradhan:2012:CST:2391181.2391183} & Media, Texts, Conversational & CoNLL\\
    ParCor \citep{ParCorDataset} & Media, Conversational & MMAX2 \\
    ParCorFull \citep{lapshinova-koltunski-etal-2018-parcorfull} & Media, Conversational & MMAX2 \\
    ECB+ \citep{ecbplus:cybulska2014guidelines} & Media & ECB+ \\
    CIC \citep{CicDataset:W16-3612} & Conversational & CoNLL \\
    WikiCoref \citep{ghaddar2016wikicoref} & Texts & CoNLL \\
    GUM \citep{GumDataset} & Texts, Conversational & CoNLL \\
    PreCo \citep{PreCoDataset}& Texts & CoNLL \\
    WinoBias \citep{zhao-etal-2018-gender}& Media, Texts, Conversational & Winograd \\
    Mind the GAP \citep{GapDataset} & Texts & Winograd \\
    KnowRef \citep{emami-etal-2019-knowref} & Media, Texts, Conversational & Winograd \\
    LitBank \citep{Bamman2020AnAD} & Texts & Brat \\
    
\end{tabular}}
\caption{Coreference Resolution corpora}
\label{tab:datasets_table}
\end{table}

As a result, the document categories that comprise each dataset are highlighted in Table \ref{tab:datasets_table}, along with the annotation format of each dataset. The CoNLL format described in \cite{hovy2006ontonotes} is used by the majority of the CR datasets after 2011. Exceptions are the ECB+ dataset which uses its own format \citep{ecbplus:cybulska2014guidelines}, the ParCor and ParCorFull which use the MMAX2 schema \citep{muller2006multi-mmax2} and the WinoBias, Knowref and Mind the GAP datasets that use Winograd style schemas \citep{levesque2012winograd}. However, Mind the GAP uses a looser format of the schema that does not contain reference-flipping words. Futhermore, the GUM dataset is also available in other formats such as ANNIS \citep{krause2014annis3}.

Effectively, all of aforementioned datasets created prior to 2017 are either very small in size like, the ParCor dataset, or aim towards a very specific Coreference Resolution tasks and are therefore an unsuitable replacement of the CoNLL 2012 as the benchmark dataset for unrestricted Coreference Resolution. Also, due to the recency of newer datasets, created post 2017, these have not been thoroughly tested and do not appear in modern research. As a result, none of them has been able to replace the CoNLL 2012, which is still being used to benchmark new CR approaches. The WikiCoref dataset represents an exception as it is being utilized for out-of-domain evaluation experiments, but does not serve as an actual replacement. {Similarly, the LitBank dataset has started to appear in the latest publications but serves as a supplementary performance evaluation dataset.}

\section{Entity Coreference Resolution evaluation metrics}
\label{sec:metrics}
This section describes in detail the three metrics used in research for the evaluation of Entity Coreference Resolution tasks models (MUC, B-cubed and CEAF), briefly touches on alternative metrics and discusses their advantages and disadvantages. These three metrics are important as they provide comparison bases with previous research. Each metric is described in terms of how it calculates the Precision and Recall. The F1-score is defined as the harmonic mean between the two in all metrics.

Within the scope of this section we use a global notation for all metrics. We denote a coreference chain as $C$ and the number of mentions in the chain as $|C|$. The term key chains refers to gold coreference chains, while system chains refers to system generated (or predicted) chains. $K(d)$ and $S(d)$ identify the set of gold coreference chains and predicted coreference chains respectively, and can be represented as:  
\begin{equation*}
    \begin{aligned}
    K(d) = \{ K_i : i = 1,2, \dots , | K(d) | \},  \\ 
    S(d) = \{ S_j : j = 1,2, \dots , | S(d) | \}, 
    \end{aligned}
\end{equation*}
where $K_i$ and $S_j$ represent chains in $K(d)$ and $S(d)$ respectively, and $|K(d)|$ and $|S(d)|$ represent the number of chains in the sets.

\subsection{MUC}
The MUC score was the first scoring metric introduced to the task of Coreference Resolution by \cite{Vilain:1995:MCS:1072399.1072405} for the 6$^{th}$ Message Understanding Conference for the task of Coreference Resolution. It identifies references as linked references, where each one can be linked to a maximum of two other references. This is achieved by counting the changes (insertions and deletions) required in the predicted (system) set to make it identical to the gold standard key set. 
In order to express Precision and Recall, we first identify a partition as: 
\begin{equation}
    P(S_j) = \{ C_j^i : i = 1,2, \dots, |K(d)|\}, \text{ where } C_{j}^{i} \text{ is } S_{j} \cap K_{i} \text{ .}
\end{equation}
Then we can use the subset $C_j^i$ to calculate the number of common links as: 
\begin{equation}
    \begin{aligned}
        c(K(d),S(d)) = \sum_{j=1}^{|S(d)|} \sum_{i=1}^{|K(d)|} w_c (C_j^i), \\
        \text{where } w_c (C_j^i) = \begin{cases}
                                    0 & \text{if }|C_j^i|=0;\\
                                    |C_j^i| - 1 & \text{if }|C_j^i|>0.
                                    \end{cases}
    \end{aligned}
\end{equation}
In Eq. \{2\} $w_c (C_j^i)$ is commonly identified as ``weight'' of $C_j^i$, which represents the minimum number of links needed to create the cluster. 
Similarly, the number of links in the key $k(K(d))$ and the number of links in the system chain $s(S(d))$ are calculated as: 
\begin{equation}
    \begin{aligned}
    k(K(d)) = \sum_{i=1}^{|K(d)|} w_k(K_i), \text{ where } w_k(K_i)=|K_i|-1 \\
    s(S(d)) = \sum_{j=1}^{|S(d)|} w_s(S_j), \text{ where } w_s(S_j)=|S_j|-1
    \end{aligned}
\end{equation}

Finally, Precision and Recall are defined using Eqs. \{2 $\And$ 3 \} as: 
\begin{equation}
    \begin{aligned}
     Precision = \frac{c(K(d),S(d))}{s(S(d))} , \\
     Recall = \frac{c(K(d),S(d))}{k(K(d))}
    \end{aligned}
\end{equation}

From this we can interpret that the MUC score can not be used to identify signleton entities (i.e. entities only mentioned once), because of its link based approach, and therefore can not be trusted to score datasets like the ACE dataset \citep{ACE:L04-1011}. 

\subsection{B-cubed}
The B-cubed is a metric introduced by \cite{Bagga1998AlgorithmsFS}, and designed to overcome the problems of MUC score. It does not take into account the links, but calculates Precision and Recall for each mention individually in the document, and uses a weighted sum to calculate the final Precision and Recall. 
With $m_n$ as the $n$-th mention in a document $d$ and $C_{j}^{i} = S_{j} \cap K_{i}$ where $S_{j}$ and $K_{i}$ are the key and system chains respectively, we define : 
\begin{equation}
    \begin{aligned}
        Precision(m_n) = \frac{w_c(C_j^i)}{w_k(K_i)}, \\
        Recall(m_n) = \frac{w_c(C_j^i)}{w_s(S_j)},
    \end{aligned}
\end{equation}
where  $w_c(C_j^i)=|C_j^i|, w_k(K_i) = |K_i| \And w_s(S_j) = |S_j|$. Then, Final Precision and Final Recall are calculated as: 
\begin{equation}
    FinalPrecision = \sum_{n=1}^{N} Precision(m_n),
\end{equation}
\begin{equation}
    FinalRecall = \sum_{n=1}^{N} Recall(m_n),
\end{equation}
where $N$ represents the number of entities in the documents.

The B-cubed implementation has a significant flaw. The approach metrics, both precision and recall, are computed by checking if the entities are containing the mention, leading to inaccurate results in cases where entities are being used more than once. 

\subsection{CEAF}
CEAF, which stands for Constrained Entity Alightment F-measure, was introduced by \cite{Luo:2005:CRP:1220575.1220579}, and was designed to fix the drawbacks of the B-cubed scoring metric. It aims to find a one-to-one mapping ($g*$) between the $K(d)$ and $S(d)$ chains using the Kuhn-Munkres algorithm \citep{kuhnh:doi:10.1002/nav.3800020109}, and a similarity measure $\phi$ to evaluate the similarity between the entities. The mapping function is defined as $g$ with a scoring function $\Phi(g)$ as: 
\begin{equation}
    \begin{aligned}
        \Phi(g) = \sum_{K_{i} \in K_{\min}(D)}\phi(K_i,g(K_i)), \\
        \text{where } g(K_{i}) = S_{j}, K_{i} \in K_{\min(d)} \\ 
        \text{ and } S_{j} \in S_{\min(d)},
    \end{aligned}
\label{eq:ceaf-scoring}
\end{equation}
with $\phi$ being the function that calculates the similarity between the gold and system chains. 
With the use of the scoring function in Eq. \{\ref{eq:ceaf-scoring}\} we can identify the optimal mapping $g*$ and with it define Precision and Recall as: 
\begin{equation}
    Precision = \frac{\Phi(g*)}{\sum_{i=1}^{|K(d)|} \phi(K_i,K_i)}
\end{equation}
and 
\begin{equation}
    Recall = \frac{\Phi(g*)}{\sum_{j=1}^{|S(d)|} \phi(S_j,S_j)}
\end{equation}
In \citep{Luo:2005:CRP:1220575.1220579}, four different similarity functions are considered: 
\begin{equation}
    \begin{aligned}
        \phi_1(K_i,S_j) = \begin{cases}
                            1 & \text{if } K_i = S_j\\
                            0 & \text{otherwise }
                            \end{cases} \\
    \end{aligned}
\end{equation}
\begin{equation}
    \begin{aligned}
        \phi_2(K_i,S_j) = \begin{cases}
                            1 & \text{if } K_i \cap S_j \neq \emptyset \\
                            0 & \text{otherwise }
                            \end{cases} \\
    \end{aligned}
\end{equation}
\begin{equation}
    \begin{aligned}
        \phi_3(K_i,S_j) = |K_i \cap S_j| = w_c(C_j^i) \\ 
    \end{aligned}
\label{eq:ceaf-phi3}
\end{equation}
\begin{equation}
    \begin{aligned}
        \phi_4(K_i,S_j) = \frac{2 * |K_i \cap S_j|}{|K_i| + |S_j|} = 
                            \frac{2 * w_c(C_j^i)}{w_k(K_i)+w_s(S_j)} 
    \end{aligned}
\label{eq:ceaf-phi4}
\end{equation}

Out of the 4 different similarity functions, only the functions described in Eqs. \{\ref{eq:ceaf-phi3} and \ref{eq:ceaf-phi4}\} are used and are considered the two variations of CEAF metric as CEAF$_m$ for mention-based CEAF and CEAF$_e$ for entity-based CEAF respectively.\\

Apart from the three metrics described in detail, which are being used in parallel in all recent research, more metrics have been introduced through the years to solve issues with these metrics, that are not covered in as much detail in this section. 
This is because of the consistent use of the CoNLL 2012 dataset and the CoNLL metric, by all researchers, that was proposed along the dataset \citep{Pradhan:2012:CST:2391181.2391183}, which is the unweighted average of the F1-scores of the MUC, B-Cubed and CEAF metrics. Another argument for the constant use of the CoNLL metric is the replicability and comparabilty of the results between works and the ability to define the state-of-the-art. Moreover, detailed studies, highlighting the advantages and disadvantages of these metrics have been conducted \citep{W10-4305,Pradhan2014MetricsPartitions,N15-3002}. 

Other metrics for Coreference Resolution include the B-cubed extensions by \cite{Stoya:P09-1074} to handle twinless mentions, and the ACE evaluation scoring \citep{ACE:L04-1011}, which was introduced to accommodate the ACE conference. In the recent years, the BLANC metric \citep{Recasens2011BLANCIT} which is a Rand-Index based metric was proposed, to solve the issue of high scores deriving from singletons in the MUC and B-Cubed metrics. 

Further metrics in the form of the LEA metric \citep{LEA:P16-1060} have been proposed, 
which take into account the importance of each entity and an entity resolution score, leading to higher scores by resolving entities with more mentions.  However, as Coreference Resolution is not an end task, the importance of resolving named entities is significant for downstream tasks \citep{chen2013linguistically}. Recently, the NEC evaluation metric was proposed \citep{agarwal-etal-2019-evaluation}, giving emphasis on resolving named entities and their importance on downstream tasks such as Entity Linking.{ What is more, }\cite{le-titov-2017-optimizing}{ proposed altering B-cubed and LEA in such a way that they are differentiable, through relaxation (i.e. using soft-clusters), in order to be used as a training objective for coreference resolvers.}

For the subtask of Pronoun Resolution, there are no special evaluation metrics introduced, since it is defined and developed as a binary classification task of different pronoun types (e.g. Third Personal, Possesive, etc.). \cite{emami-etal-2019-knowref} proposed the use of a Consistency score. While this score does not replace the evaluation metrics of the task, by swapping the entities in the test set, it enables examination of the role of contextual information in the system. 
Gendered Pronoun Resolution is treated as a multi-class classification problem amongst Masculine, Feminine and Neither, and the results for each class are calculated by conventional multi-class Precision and Recall metrics.

\section{The evolution of Entity Coreference Resolution}
\label{sec:CRhistory}
Through the many years of research, CR has been approached with four different techniques. This attests to the difficulty of the problem due to the fact that the techniques used were built on top of each other in a hierarchical way.

We identify the following approaches deployed to solve CR with chronological order of appearance: Mention-Pair models, Mention-Ranking models, Entity-Based models, Latent Structured models.

Mention-Pair models form the simplest and purest form attempted in CR, examining a pair of mentions at a time, along with the features of each mention, and assigning a binary outcome \citep{soon2001machine, ng2002identifying, denis2007joint}. 

Mention-Ranking models come to solve the most obvious disadvantage of the Mention-Pair models, not considering dependency with the other candidate antecedents, by simultaneous ranking and making a connection only with the highest ranking antecedent \citep{yang2003coreference, rahman2009supervised}. 

While Mention-Ranking models set the new state-of-the-art at the time, they lacked the ability to determine when clusters should not be merged. As transitivity plays a big role in CR, without the ability to take it into account, decisions to pair mention clusters together would inevitably yield mistakes. Entity-Based models offered a means to classify the knowledge required to enable informed merge decisions \citep{luo2004mention, yangy2004np, ratinov2012learning, stoyanov2012easy}. This was implemented in Entity-Mention models and Cluster-Mention models, with the latter showing significant improvements. 

Following on the steps of making previous models that attempted to map entities, Latent-Structure models have made an appearance, shifting the focus from creating agglomerative clustering iteratively to create a tree-like structure that coreference partitions can be extracted from it \citep{fernandes2012latent, durrett2013easy, bjorkelund2014learning}. 

In the recent years, most applications began traversing from simple machine learning techniques to deep learning. This is due to a combination of upgrades in hardware capabilities that enable complex neural models, as well as advancements in architectures and methodologies that are capable to both generalize better and effectively utilize vast data. A significant milestone that signalled the start of the DL era is the concept of word embeddings \citep{mikolov2013distributed} followed by more advanced language representation methodologies \citep{Peters:2018,DBLP:journals/corr/abs-1810-04805}.

We notice that essentially, Mention-Pair and Mention-Ranking models set the foundation required for Entity-Based and Latent-Structure models as they represent core components for their functionality. A detailed review of the approaches described above can be found in \cite{Ng:2010:SNP:1858681.1858823}.

However, these advancements have also introduced new challenges in the form of gender bias. \cite{kurita2019measuring} provides a comprehensive methodology into measuring gender bias in ELMo, while \cite{zhao2019gender} also discusses the error and provides two different methodologies (i.e. data augmentation and neutralization) to mitigate the bias in the representations. As a result, Pronoun Resolution, a subtask of Entity CR, has been distinctly identified and tackled, to solve the bias in CR resolvers. We make a distinction between the approaches of the two tasks, with the former describing a clustering problem while the latter describing a binary classification problem. 

In the following sections we shall analyze the advances introduced in the DL approaches, as well as provide an overview of the Pronoun Resolution subtask. We notice a similar trend between deep learning CR and conventional CR approaches, starting from Mention-Pair type models and gradually moving to Entity-Based and Latent-Structure approaches, iteratively setting the state-of-the-art bar higher. On the other hand, Pronoun Resolution, as it is a newly tackled task, does not see the same evolutionary lineage, and the majority of the novelties lie to de-biasing methods. 

\section{A review of Entity Coreference Resolution methodologies}
\label{sec:ECR}
At the first stages of DL approaches, the mentions from the documents were extracted using either the Berkeley Coreference System (BCS) \citep{durrett2013easy} or the Stanford Deterministic Coref System rules \citep{lee2011stanford}, while the latter was used to extract animacy features in implementations up to 2017. At early 2017, the first end-to-end CR systems made their appearance. As a result, the implementations are very different, since the input transitions from mentions to spans of text. 

Following the methodologies used in early CR, neural approaches can be identified in similar categories. We identify five major categories for neural CR: Mention-Pair models, Mention-Ranking models, Entity-Based models, Latent-Structure models and Language-Modeling models. We futher categorize Latent-Structure models and Language-Modeling models as Improvement model types, due to the fact that the described approaches are heavily based on either Mention-Ranking models or Entity-Based models. We notice that deep learning coreference systems follow the same iterative approach used to build non-neural coreference systems. Mention-pair models are adaptations of their non-neural counterparts using simple perceptron architectures. Due to the many downfalls of such approaches, the neural models quickly evolved to Mention-Ranking models.
Moreover, as Mention-Ranking models are at the core of Entity-based models that utilize the scoring functions to prune possible antecedents, implementations of the first were central to implementation of the second. Neural networks have however enabled the effective implementation of Latent-structure models with a combination of graphs and clusters, further extending past approaches. Finally, approaches that excelled in CR have been identified through Language-Modeling models which are not directly aimed for CR. We analyze each of those approaches separately in the sections below. 

\subsection{Mention-Pair models}
Neural Mention-Pair approaches, were only attempted in the very early ages of neural networks. As a result, they are focused on direct implementations of past methodologies using perceptrons as their core component and enhancing the feature space to achieve state-of-the-art results.

The first neural approach to solve CR, introduced by \cite{bengtson2008understanding} is an average perceptron  pairwise model, following the work of \cite{ng2002identifying}, which implements a best-first clustering. While the neural approach is trivial, the work is focused on advanced feature selection and their importance in the task. An extensive use of WordNet \citep{fellbaum1998semantic} for the extraction of semantic features and their incorporation is the authors` main contribution.

Similarly, \cite{charton2011poly} have introduced their own approach, in which they implement a multi-layer perceptron in conjunction with other non neural approaches. They implement a closest-first clustering adaptation \citep{soon2001machine}, while the methodology's main contribution lies in the use of novel features, such as named entity clustering and alias searching, as well as similarity metrics introduced to incorporate these features for a pair of candidates. However, even though the neural model did not outperform the non-neural approaches developed by the authors with the same set of features, there was no measurable statistical difference between the neural and non-neural approaches. 

\subsection{Mention-Ranking models}
\cite{wiseman2015learning} proposed the first mention-ranking approach to solve CR, which expands the mention-ranking scoring function described in \cite{chang-2013-constrained} to a piece-wise scoring function that distinguishes between the mention being anaphoric or not. 

They define their neural network model as an adaptation of the piece-wise scoring function via a feed-forward neural network. It makes use of both the BCS to extract mentions and the Stanford Deterministic Coref System rules to extract two sets of features, BASIC \citep{durrett2013easy} and BASIC+, which extends them with features utilized in \cite{recasens-2013-life}, and create the feature representations. However, in comparison to past approaches, they use raw, unconjoined features extracted via BCS and pretrain on the subtasks of anaphoricity and antecedent ranking to intialize the weights of the feature representations, before training directly on CR. The model is trained to minimize the regularized, slack-rescaled, latent-variable loss defined as: 
\begin{equation}
    L_{p}(\theta_{p}) = \sum_{n=1}^N \max_{\hat{y} \in Y(x_{n})} \Delta_{p}(x_n , \hat{y})(1 + s(x_{n}, \hat{y})-s(x_{n}, y_{n}^{\ell})) + \lambda\|\theta\|_{1} \texttt{ ,}
\label{eq:wiseman-loss-function}
\end{equation}
which makes use of a mistake-specific cost function $\Delta$ of three mistakes, namely ``false link'' (FL), ``false new'' (FN), ``wrong link'' (WL) that are individually defined. The subtasks use directly analogous loss functions for the pretrained tasks too. 

\begin{figure}[!ht]
    \centering
    \includegraphics[width=\textwidth]{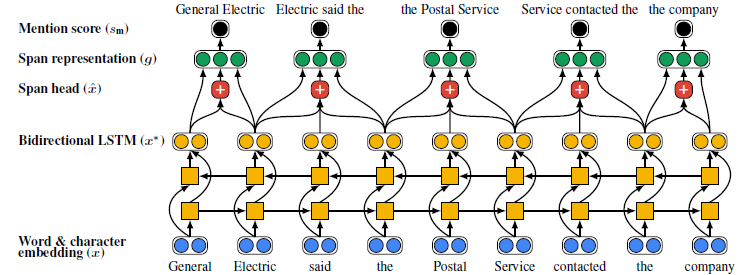}
    \caption{End-to-end Coreference resolution model architecture from \cite{lee2017end}.}
    \label{fig:lee_model}
\end{figure}

Shifting from mention parsers and tools to extracting mentions, in \cite{lee2017end}, the first end-to-end CR approach is described, which instead of mentions, considers all possible spans of text and learns to identify mention spans and how to pair them into clusters.{ Figure }\ref{fig:lee_model} {presents an overview of the model architecture used in the end-to-end model.} The model uses a pairwise scoring function that, for each pair of spans in question, takes into account a unary mention score of each span and a pairwise score of the two spans in question.

\begin{figure}[!ht]
    \centering
    \includegraphics[scale=0.8]{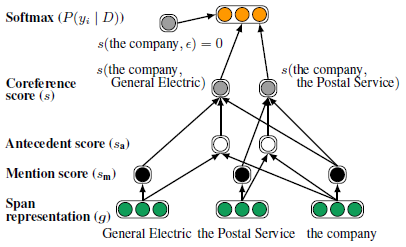}
    \caption{Antecedent scoring component from \cite{lee2017end}.}
    \label{fig:lee_scoring}
\end{figure}

{The scoring functions are implemented via a two-layer feed-forward neural network (FFNN), illustrated in Figure }\ref{fig:lee_scoring}, while the span representations are being computed using bidirectional long short-term memory networks (BiLSTMs) to capture lexical information from the whole text. The system also uses an attention mechanism \citep{bahdanau2014neural-attention} to identify the head words in span representations. As the antecedents are not predefined, the loss of the system is defined as:
\begin{equation}
\log \prod_{i=1}^{N} \sum_{\hat{y} \in Y(i) \cap GOLD(i)} P(\hat{y})
\label{eq:lee-loss-function}
\end{equation} 
since it uses only the gold mention clusters.  However, the system has to apply significant hard limitation to span sizes and distance between the spans and candidate spans. The latter is done by constraining spans based on the scores produced through the scoring function. 

In \cite{zhang-etal-2018-neural-coreference-biaffine}, a different antecedent scoring mechanism is proposed, using the end-to-end span ranking system described in \cite{lee2017end} as the baseline. With the use of biaffine attention to calculate clustering scores, it allows the scoring function to directly model the compatibility of two mentions and the prior likelihood of being connected. Furthermore, the system extends the loss function (Eq. \{\ref{eq:lee-loss-function}\}) to optimize not only the clustering, but clustering and mention detection performance jointly, making use of the biaffine scoring mechanism.

To attempt and solve the common problem of globally inconsistent decisions amongst Mention-Ranking models, without using global features or entities, a clustering algorithm is proposed in \cite{gu2018study}. Using the span ranking model as a baseline \citep{lee2017end}, they propose the use of indirect links via the scoring function and create sets that are considered during inference to dismiss clustering decisions.

\subsection{Entity-Based models}
Neural Entity-Based models are adaptations or extensions of the previous Mention-Ranking models that attempt to incorporate global features with different methodologies. At the core of Entity-Based approaches, they are using an adaptation of the Mention-Ranking functions to allow for Cluster-Mention ranking.

In \cite{wiseman2016learning}, a Cluster-Based approach is proposed, with clusters holding features of the individual mentions separately, in a global representation. The cluster features are then used in a global scoring function between the cluster assigned to each possible antecedent and the mention, along with a local scoring function of the mention and each possible antecedent.  
The systems uses long short-term memory (LSTM) networks, to embed cluster features and the same representation for both individual mentions and mention-pairs, as in \cite{wiseman2015learning}. The states of the RNNs before a decision is made for the current mention are utilized in the global scoring function to make an effective map of previous decisions. As the model is trained directly on the task of CR, the slack-rescaled loss function described in Eq. \{\ref{eq:wiseman-loss-function}\} is optimized to reflect cluster information, while adopting the same mistake-specific cost function. 

A similar approach has been used in \cite{clark-manning-2016-improving} where they also use clusters to capture global information, and cluster ranking to make merging decisions. In comparison to \cite{wiseman2016learning}, they define each mention as a single entity cluster and combine them during inference. The system is comprised of three different components that work together to feed the respective representations to a single-layer cluster ranking model that makes merging decisions.

\begin{figure}
  \centering
    \begin{subfigure}[t]{0.45\textwidth}
        \centering
        \includegraphics[width=\linewidth]{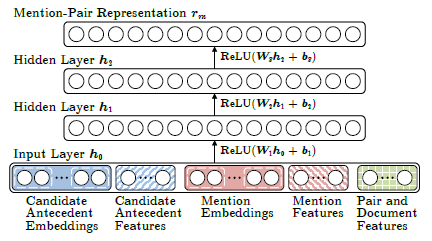}
        \caption{Mention-pair encoder} \label{fig:clark_mention_pair}
    \end{subfigure}
    \hfill
    \begin{subfigure}[t]{0.45\textwidth}
        \centering
        \includegraphics[width=\linewidth]{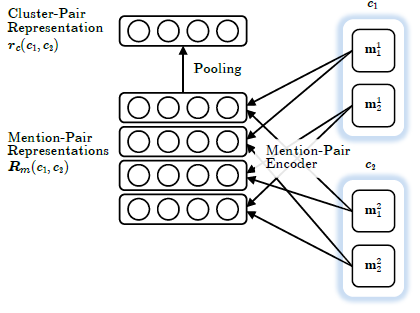}
        \caption{Cluster-pair encoder} \label{fig:clark_cluster_pair}
    \end{subfigure}
    \caption{Mention-pair encoder and Cluster-pair encoder from \cite{clark-manning-2016-improving}.}
\end{figure}

{The mention-pair encoder, as illustrated in Figure }\ref{fig:clark_mention_pair}{, is a three-layer fully-connected FFNN with a Rectified Linear Unit (ReLU) activation }\citep{Nair:2010:RLU:3104322.3104425}{ on each layer that takes the embeddings, the features of each mention and the candidate antecedent, in order to produce a high-level representation. The cluster-pair encoder (Figure }\ref{fig:clark_cluster_pair}{) produces a distributed representation for a pair of clusters by applying max pooling and average pooling, and concatenating the results for all mention-pairs in each clusters.} The mention-ranking model is an adaptation of \cite{wiseman2015learning}, that scores all mention pairs produced by the mention-pair encoder. It uses the same loss function described in Eq. \{\ref{eq:wiseman-loss-function}\} with the same mistake-specific function. However, pretraining is performed based on the two objectives described in \cite{clark-manning-2015-entity} (All-Pairs Classification and Top-Pairs Classification). 

Finally, the Cluster-Ranking model is a single-layer network that utilizes the cluster-pair encoder and an anaphorocity score to assign cluster scores. The decision to combine (MERGE) or not (PASS) the cluster is made by policy network $\pi$ that takes into account cluster ranking and anaphoricity scores of the cluster ranking model. However, as future decisions are based on previous ones, the system is utilizing a learn-to-search algorithm \citep{chang2015learning} to project all possible actions taken by the policy network and is trained to minimize the risk associated with each action, while also sorting mentions in a descending order using the scores produced by the mention-ranking process. 

Because the policy network is very hard to optimize correctly and is also directly related with the coreference evaluation metric B-cubed, an extension is decribed in \cite{clark-manning-2016-deep} using Reinforcement Learning (RL). Specifically, two different RL algorithms are used to replace the learn-to-search approach that optimizes the merging policy. This is possible due to the distinct actions of the policy that can be translated into an action-reward system. The Reward Rescaling and REINFORCE algorithms described are attempting two different approaches to optimize the policy. The first trains the agent to be able to map the reward of each individual action and the second attempts to maximize the expected reward by calculating the probabilistic distribution of an action using the mention-ranking model. Both approaches outperformed the learn-to-search approach in different metrics. 

Building on top of \cite{clark-manning-2016-improving}, the Sanaphor++ system is proposed 
\citep{plu2018sanaphor++}. It utilizes semantic knowledge from external data source, specifically Wikipedia, and ontologies from DBPedia and YAGO. The system applies ADEL and Sanaphor \citep{plu2016knowledge, prokofyev2015sanaphor} to provide entity links to mentions, the entities of which are known, and utilize the NER tags for entities that are unknown, in order to provide knowledge. The annotated information is used along with the mentions of the system and utilized by an optimized mention-pair ranking model that takes into account the entity information by optimizing the mistake-specific function during training (Eq. \{\ref{eq:wiseman-loss-function}\}). 

{In an attempt to limit the memory requirements of incrementally building clusters }\cite{xia2020revisiting}{ and }\cite{toshniwal2020learning}{ have both introduced neural approaches that are extensions of the work done by }\cite{webster2014limited}{. Both approaches are using }\cite{joshi2019spanbert}{ as their baseline with pre-training for weight initialization, in order to encode the entity information in the document and acquire scores on candidate antecedents. }\cite{xia2020revisiting}{ uses a similar approach to }\cite{lee2017end}{
during training, with the alteration that it only considers the most recent antecedent in an entity cluster for cluster matching and updates gradients once per document and not once per mention. During inference, to incrementally build the clusters while maintaining a constant space, it prunes the entity space based on entity cluster size and distance from the current span considered.}

\cite{toshniwal2020learning}{ builds on top of }\cite{xia2020revisiting}{, however instead of using naive cluster pruning, they define three measures to be used for a learned bounded memory and a rule-based bounded memory approaches. These measures are, the number of tokens between the first and last entity mention (entity spread), the number of entity spreads that include a specified token (active entity count) and the maximum number of active tokens at any given token in the document (maximum active entity count). They also offer an unbound memory approach which differs from }\cite{xia2020revisiting}{ in that they don't append non-coreferent mentions. The learned bound memory approach utilizes a learned score to predict the number of remaining entity mentions and prunes clusters based on that decision while the rule-based approach is pruning the least recently used entity cluster.}

\subsection{Latent-Structure models}

An extension of the work in \cite{lee2017end} is described in \cite{lee2018higher}, which improves the baseline on two aspects. First, it allows for a refined span representation, through iterations, with the use of a gated attention mechanism. As a result, the span-ranking mechanism is predicting latent antecedent trees, with each parent of a span and each tree representing a cluster. Secondly, as the complexity is increasing dramatically on long documents, an antecedent pruning mechanism is applied that is based on an altered scoring mechanism. The altered scoring function uses a less accurate and less computationally expensive score for antecedent and cluster score functions that are leveraged in a three-stage beam search during final inference. This, along with the use of the refined spans, effectively alleviates the need for a priori knowledge and distance heuristics of antecedents in a document, which was used in the baseline model. 

In an approach aiming towards better generalization, an adverserial training technique has been proposed in \cite{subramanian2019improving}. The system, which is based on the model of \cite{lee2018higher}, calculates a loss gradient for each span based on its representation, and modifies the loss function to take into account the adversarial loss using the fast-gradient-sign-method \citep{miyato2016adversarial}. Focusing on the named-entity problem, described by \cite{chen2013linguistically}, they create adverserial examples by replacing named-entities in the test set and ensuring that no data overlap exists.

As Coreference Resolution is heavily related to World Knowledge, \cite{aralikatte-etal-2019-rewarding} describe a system that makes use of Relation Extraction systems and the ``distill'' multi-task reinforcement learning technique described in \citep{teh2017distral} to create a reward system for coreference resolvers. The coreference resolution system is based on \cite{lee2018higher} and uses policy gradient and model interpolation based on a reward produced by three different relation extraction systems. Each relation extraction system captures a different relation aspect and calculates a reward based on that aspect, with the final reward being the normalized sum of individual rewards. 

{Further work has been done on expanding the entity representations in }\cite{lee2018higher}{ through Entity Equalization in }\cite{kantor2019coreference}{. This is achieved by changing the span representations of the antecedents in order to contain information about their respective entity cluster. To do so, they create an entity representation for each entity at each timestep and then calculate the entity distributions of each mention with their global entity representations. Furthermore, as this process required differentiable entity cluster representables in order to be able to train end-to-end, they adopt a soft-clustering approach from }\cite{le-titov-2017-optimizing}{. Their approach replaced ELMo with BERT embeddings and also remove the second-order span-representations as they noted better overall performance.}

{Based on the issues highlighted by }\cite{kantor2019coreference}{, }\cite{xu2020revealing}{ performed an extensive analysis on higher-order inference when using SpanBERT for their representations. They present two higher-order inference clustering approaches, span clustering and cluster merging, and compare them with the approaches described in }\cite{lee2018higher}{ and }\cite{kantor2019coreference}{. Span clustering is similar with the soft-clustering approach in }\cite{kantor2019coreference}{however it constructs the actual clusters instead of calculating the probabilities, while cluster merging utilizes antecedent and cluster information for antecedent reranking, in comparison to }\cite{lee2018higher}{. They verify the previously mentioned higher-order inference issue and conclude that cluster merging performs best.}

{In the same time, }\cite{liu2020improving}{ explored the use of Graph Neural Networks instead of cluster, basing his approach on }\cite{lee2018higher}{, with BERT embeddings instead of the ELMo used in the baseline. The proposed model uses graphs to model entities, with nodes being the entity mentions and the edges refer to the weight of the neighbouring nodes to their connected nodes. These entity information, in the form of edge weights, are taken into account when calculating the antecedent scoring.}

\begin{figure}[!ht]
    \centering
    \includegraphics[scale=0.7]{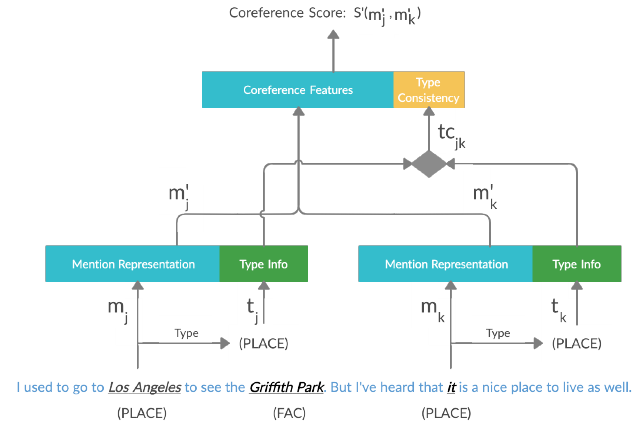}
    \caption{Type information for improved CR as presented in \cite{khosla2020using}.}
    \label{fig:khosla_types}
\end{figure}

\cite{khosla2020using}{ describes an extension of the work done in }\cite{lee2017end}{ with the exception of using BERT embeddings and only focusing on the task of mention linking }\citep{Bamman2020AnAD}{. In this extension, type information in the form of named entity types in the case of CoNLL 2012 (Figure }\ref{fig:khosla_types}{), are utilized to reduce type inconsistency in the predicted clusters. This is achieved through the concatenation of the type information in the mention representations and through the use of the type information in the feature vector for the scoring function.}

\begin{figure}[!ht]
    \centering
    \includegraphics[width=\textwidth]{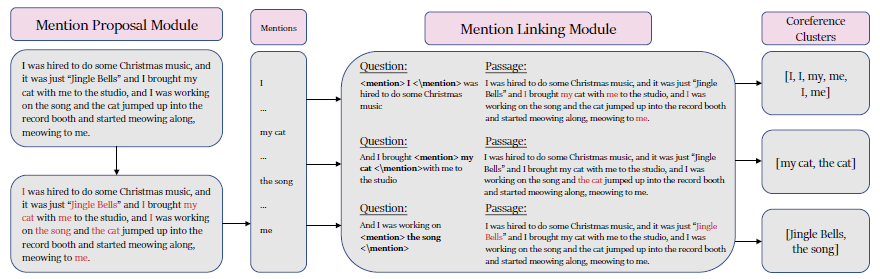}
    \caption{CorefQA model pipeline from \cite{wu-etal-2020-corefqa}.}
    \label{fig:wu_corefqa}
\end{figure}

\cite{wu-etal-2020-corefqa}{ establishes a novel approach to handle Coreference Resolution as a Question-Answering (QA) task. By changing the formulation to a QA task, the resulting model is able to re-use disregarded mentions when they are considered unfit which are unrecoverable to the vast majority of the other implementations and also to incorporate contextual information when linking mentions through query generation. The model is comprised of three components (Figure }\ref{fig:wu_corefqa}{), the Mention Proposal module, the Mention Linking module and the Coreference Clustering.} 

{The Mention Proposal module operates similarly to }\cite{lee2017end}{ computing a mention score for all candidate mentions in the span and performing a greedy pruning based on the scores. In comparison to the base approach, it uses SpanBERT for span representations while also implements the Overlap segmentation approach from }\cite{joshi-etal-2019-bert}{. Furthermore, instead of the feature-based baseline approach, the model explicitly uses the speaker information by concatenating the speaker name to the respective text.}

{The Mention Linking module calculates the antecedent scoring function via a question answering framework using the \{Context, Query, Answers\} triplet as input. As a result, it considers the whole document as Context, which allows the model to incorporate more contextual information. Each query is constructed using the sentence that the mention is identified in, annotated with special tokens for the model to be able to locate it. The answers are all the coreferring mentions in the document. The Context and Query are used as a single sequence in the model while the answers are provided in a BIO (Before, Inside, Outside) format though a softmax function, to enable multiple predictions which is part of the model's ability to reconsider previously unfit mentions. Furthermore, the scoring mechanism is extended to bidirectional mention scoring, which is the average score of the respective mention scores when given two mentions with each mention being the antecedent and the other being the mention. The module is trained using the same training objective as in }\cite{lee2017end}{ (Eq. }\ref{eq:lee-loss-function}{), using the bidirectional scoring, while it is also pre-trained on question answering for data augmentation.} 

\subsection{Language-Modelling models}
We refer to Language-Modelling models for CR to models that are not built to solve CR, but to represent tasks of modelling linguistic phenomena that are thereafter used to enhance CR and other NLP tasks. As a result, the majority of the models have been implemented and tested with fine-tuning on CR tasks. 

A way to use syntactic information, in a treebank format, to infuse spans of text is proposed in \cite{swayamdipta-etal-2018-syntactic}. While using multi-task learning to create syntactic information, the system is unrestrained into creating valid parse trees and does not assume the use of overlapping datasets for the syntactic learner and the primary task. {The syntactic scaffolds are only used to bias the decisions of the primary task, which is done by jointly learning to minimize the loss of both tasks simultaneously using a tuneable parameter }$\delta${(Eq. }\ref{eq:swayamdipta-joint-loss}{)}.{The} $L_{1}$ {and }$L_{2}${ are the syntactic scaffold and task specific losses with }$y${ and }$z${ being the respective target labels, while }$x${ represents the input.}
\begin{equation}
\sum_{(x,y) \in D_{1}} L_{1}(x,y) + \delta \sum_{(x,z) \in D_{2}} L_{2}(x,z)
\label{eq:swayamdipta-joint-loss}
\end{equation}
As a result the syntactically-aware span representations can be directly used during inference on any end task that uses spans representantions. In the case of CR, the synactic scaffolds are applied on the implementation of \cite{lee2017end}. {By enhancing span representations with syntactic information and altering the loss function described in Eq.}\{\ref{eq:lee-loss-function}\}{ to take into account the treebank format and enable joint learning (}$L_{2}${ in Eq. }\{\ref{eq:swayamdipta-joint-loss}\}{), the system boosted the performance of the baseline.} What is more, syntactic scaffolds could be implemented to further push the performance of other implementations \citep{lee2018higher,zhang-etal-2018-neural-coreference-biaffine}. 

Following the same idea, the effective use of linguistic features in early neural approaches \citep{clark-manning-2016-deep, clark-manning-2016-improving, clark-manning-2015-entity, lee2017end} is questioned in \cite{moosavi2018using}. It builds on the generalization problem due to high mention overlap in the CoNLL dataset train, development and test splits \citep{moosavi-strube-2017-lexical}. Alternatively, the authors propose a discriminative mining algorithm that mines patterns in string-matching, syntactic, shallow semantic and discource features using Frequent-Pattern Tree structure to represent them. The mining algorithm then prunes the tree by measuring the discriminative power, information novelty and frequency of the patterns that arise and enhances the results of base approaches.

Further attempts have been made for the model to consider information that exceeds the spans. This was implemented either by using cross-sentence dependency in word representations or considering more than two mentions at a ti and performing a greedy pruning based on the scoresme.

In the case of cross-sentence dependency, two different approaches were proposed in \cite{luo2018learning-crosssentence}, which are based on the \cite{lee2017end} system and improves on span representations. First, the linear sentence linking (LSL) uses BiLSTMs to intitialize the states of the BiLSTM of the second sentence using the hidden states of the first sentence, to achieve information cross. Second, they proposed an attention sentence linking (ASL) methodology in which they infuse the memory modules of LSTMs with attention, based on all previous words of the previous sentence, and apply a gated selection mechanism. As these implementations are both targeted towards changing span representations in the original systems, the rest of the model is as described in \cite{lee2017end}.

Another way to model mutual dependencies between mentions is described in \cite{meng-rumshisky-2018-triad}. Moving from dyad systems (consider only two mentions or two spans of texts at a time) to triad systems that consider three mentions at a time. As this is not an end-to-end system, the mentions are the gold mentions of the dataset and the distance between mentions as well as a binary feature to indicate speaker information are applied. The proposed methodology is described by two models, one that computes mutual dependency between the triads and one that performs the clustering. The first system uses LSTMs to represent the words and the Part-of-Speech (POS) tags to create a mention-pair representation for each pair. The triad representation is created by FFNN applying element-wise vector summations on pair representations and a decoder function that is used to measure if the representations belong to the same entity. The second system used the probability scores created by the decoders over a pair and calculates the average of their scores in all triads to make a clustering decisions.

{A direct adaptation of the }\cite{lee2018higher}{ approach, with the use of BERT to word representations is presented by }\cite{joshi-etal-2019-bert}.{ The span representations are created by using the first and last word-pieces in the span, concatenated with the attended word-pieces of all tokens in the span. Furthermore, two approaches are presented to handle span representations based on the max sequence length limitation of BERT}\citep{DBLP:journals/corr/abs-1810-04805}{, Independent and Overlap. Independent refers to the segments being continuous with no overlapping, while Overlap considers half of the previous segment as context in the next with the final token representations being derived by an element-wise interpolation of representations from both overlapping segments. While, Independent was found to perform better than Overlap, both approaches struggled with the maximum segment length of BERT, while entities were found to span over that segment size.}

\begin{figure}[!ht]
    \centering
    \includegraphics[width=\textwidth]{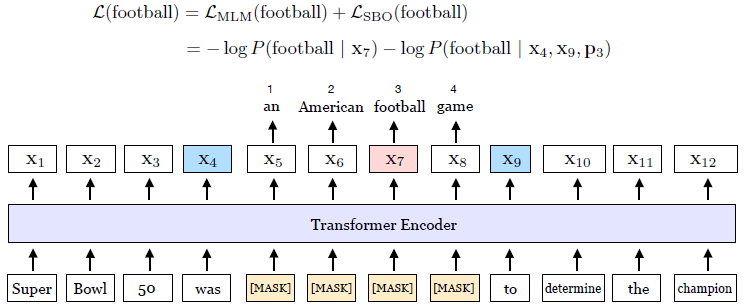}
    \caption{SpanBERT model training from \cite{joshi2019spanbert}}
    \label{fig:joshi_spanbert}
\end{figure}

{In order to resolve these weakness, as well as to increase the maximum segment limitations of BERT }\cite{joshi2019spanbert}{ introduced a modified version of the BERT model. SpanBERT alters the objectives of BERT to create span representations (Figure }\ref{fig:joshi_spanbert}{). Specifically, the token masking is altered to span masking, while the next sentence prediction is replaced with span boundary detection. In turn, this also allows for longer sequences to be models than single sentences. SpanBERT was used as a direct replacement for the span representations in }\cite{lee2018higher}{.}

\section{A review of Pronoun Resolution methodologies}
\label{sec:PR}
Pronoun Resolution has been a part of Entity CR and was therefore not handled individually until recently. However, the importance of resolving pronouns for downstream tasks along with gender bias, both within the CoNLL dataset and sentence representation techniques have made it popular. Specifically, it was noticed that less than 26\% of gendered pronouns in the CoNLL dataset are Feminine, leading to a heavy bias \citep{zhao-etal-2018-gender}. In comparison to the more difficult problem of Entity Coreference Resolution, the majority of the approaches attempting to solve Pronoun Resolution are not following clustering techniques. The approaches employed for this task are either span-ranking or binary classification (in the case of gender specific pronoun resolution). 

\begin{figure}[ht]
  \centering
    \begin{subfigure}[t]{0.49\textwidth}
        \centering
        \includegraphics[width=\linewidth]{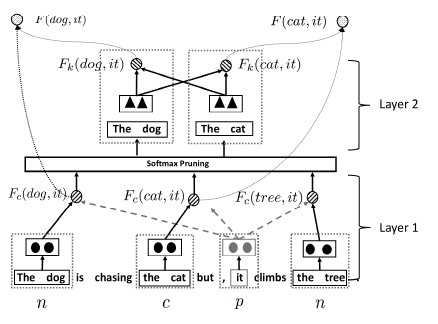}
        \caption{Pronoun resolution model \citep{zhang2019incorporating}} \label{fig:zhao_attn}
    \end{subfigure}
    \hfill
    \begin{subfigure}[t]{0.49\textwidth}
        \centering
        \includegraphics[width=\linewidth]{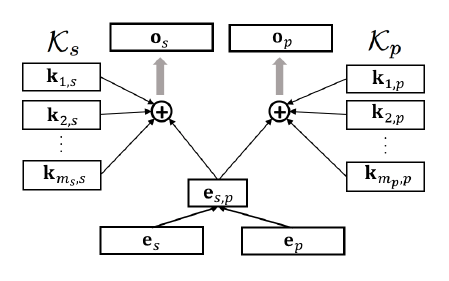}
        \caption{Knowledge attention mechanism \citep{zhang2019knowledge}} \label{fig:zhao_gk}
    \end{subfigure}
    \caption{Different approaches of incorporating external knowledge from (a) \cite{zhang2019incorporating} and (b) \cite{zhang2019knowledge}}
\end{figure}

Two different approaches to incorporate knowledge have been implemented in \cite{zhang2019incorporating,zhang2019knowledge} for pronoun resolution. Both approaches are built with a span representation approach based on the methodology described by \cite{lee2017end} and applying an inter-span attention before computing the final input representation. \cite{zhang2019incorporating}{ uses a simple FFNN and Softmax pruning to remove complexity before applying a knowledge attention mechanism, as illustrated in Figure }\ref{fig:zhao_attn}{. The attention mechanism uses external knowledge sources to weight and score the pruned inputs, resulting in the highest scored pair to be selected. }\cite{zhang2019knowledge}{ uses the same baseline, but instead of the knowledge attention mechanism, it uses knowledge graphs in the form of triplets to create a knowledge representation (Figure }\ref{fig:zhao_gk}){ for each mention and calculate a score for that pronoun pair.}

In a different approach \cite{tenney2019you} utilizes sentence representation models such as ELMo \citep{Peters:2018} and BERT \citep{DBLP:journals/corr/abs-1810-04805} with a two-layer MultiLayer Perceptron (MLP) and a sigmoid activation function. The model, although simplistic, utilizes the contextual information from span representations which are calculated using the attention pooling mechanism in \cite{lee2017end}. The methodology, described as Edge Probing, proved that sentence representation models hold contextual information outside the sentence that is leveraged towards informed decisions. 

At the same time, an in-depth study of BERT has been conducted by \cite{clark2019does}, which showcases the distinct linguistic phenomena that are represented by the attention heads and evaluates them in classifying syntactic relations. As a result of attention probing on the attention heads, it is proven that, given a coreferent mention, BERT can predict a correct antecedent. 

\begin{figure}[!ht]
  \centering
    \begin{subfigure}[t]{0.49\textwidth}
        \centering
        \includegraphics[width=\linewidth]{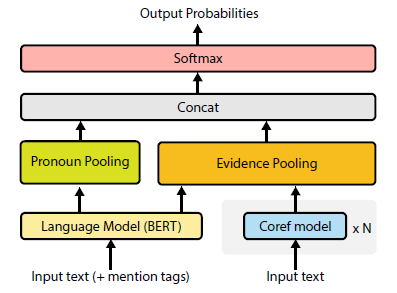}
        \caption{End-to-end model architecture \citep{attree2019gendered}} \label{fig:attree_model}
    \end{subfigure}
    \hfill
    \begin{subfigure}[t]{0.49\textwidth}
        \centering
        \includegraphics[width=\linewidth]{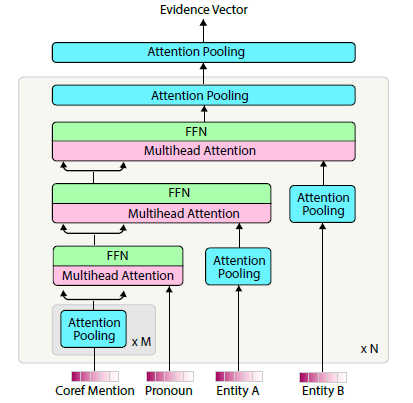}
        \caption{Coreference cluster pooler \citep{attree2019gendered}} \label{fig:attree_pooler}
    \end{subfigure}
    \caption{\cite{attree2019gendered} model architecture and Coreference pooler.}
\end{figure}

A series of methodologies have been deployed as part of the Kaggle Gendered Pronoun Resolution task. However, despite that a lot of these approaches stem from past implementations, they have managed to introduce innovations. All of the approaches use ensembles of models or predictions from entity CR systems \citep{lee2017end,GapDataset,lee2018higher} and BERT for sentence representation \citep{tenney2019you}. {The winning approach of the competition }\citep{attree2019gendered}{ introduced novelty in creating an attention pooling mechanism which uses predicted clusters of pretrained CR models (Figure }\ref{fig:attree_pooler}{) along with a fine-tuned version of BERT and a pronoun pooling methodology (Figure }\ref{fig:attree_model}{). The pronoun pooling is described as the same attention pooling operation used in the coreference cluster pooler for entities, on the annotated pronouns of the input data.}

Instead, \cite{ionita2019resolving} extracted BERT embeddings from specific layers of BERT and used an array of coreference predictions and hand-crafted features on their implementation. \cite{liu2019anonymized} introduced a data augmentation technique in which he replaced all names in the dataset to inject anonymity and make the model less biased towards the names themselves. Finally, \cite{xu2019look} did not use any external predictions, but used the BERT representations and applied a Recurrent Graph Convolutional Network architecture to capture syntax from the embeddings.

\section{Coreference Resolution Performance}
\label{sec:results}
In this section we present the best results achieved by the methodologies described in sections \ref{sec:ECR} \& \ref{sec:PR}, discuss the improvements on performance on the approaches, and set the state-of-the-art on CR to this date. {The reported scores are sourced directly from bibliography with the exception of some cases which direct contact with the authors was required as not all scores were reported. These scores are accompanied by an appropriate footnote to help distinguish them from the publications that did report all the results.}\\

\subsection{Entity Coreference Resolution results}

\begin{table}[!ht]
\centering
\caption{Neural Entity Coreference Resolution Results} 
\resizebox{\textwidth}{!}{
\begin{tabular}{l|l|lll|lll|lll|c}
\multicolumn{1}{c|}{\multirow{2}{*}{\begin{tabular}[c]{@{}c@{}}Model\\ Types\end{tabular}}} & \multicolumn{1}{c|}{\multirow{2}{*}{Models}} & \multicolumn{3}{c|}{MUC}                                                     & \multicolumn{3}{c|}{B}                                                       & \multicolumn{3}{c|}{CEAF}                                                    & \multicolumn{1}{c}{CoNLL}   \\
\multicolumn{1}{c|}{}                                                                       & \multicolumn{1}{c|}{}                        & \multicolumn{1}{c}{Prec} & \multicolumn{1}{c}{Rec} & \multicolumn{1}{c|}{F1} & \multicolumn{1}{c}{Prec} & \multicolumn{1}{c}{Rec} & \multicolumn{1}{c|}{F1} & \multicolumn{1}{c}{Prec} & \multicolumn{1}{c}{Rec} & \multicolumn{1}{c|}{F1} & \multicolumn{1}{c}{Avg. F1} \\ \hline
\multirow{6}{*}{\rotatebox[origin=c]{90}{\parbox[c]{2cm}{\centering Language modelling}}} 
    & \cite{joshi2019spanbert} & \underline{85.8}\% & \underline{84.8}\% & \underline{85.3}\% & \underline{78.3}\% & \underline{77.9}\% & \underline{78.1}\% & \underline{76.4}\% & \underline{74.2}\% & \underline{75.3}\% & \underline{79.6}\% \\
    & \cite{joshi-etal-2019-bert} & 84.7\% & 82.4\% & 83.5\% & 76.5\% & 74.0\% & 75.3\% & 74.1\% & 69.8\% & 71.9\% & 76.9\% \\
    & \cite{luo2018learning-crosssentence} & 79.2\% & 73.7\% & 76.4\% & 69.4\% & 62.1\% & 65.6\% & 64.0\% & 58.9\% & 61.4\% & 67.8\% \\
    & \cite{swayamdipta-etal-2018-syntactic} & 78.4\% & 74.3\% & 76.3\% & 68.7\% & 62.9\% & 65.7\% & 62.9\% & 60.2\% & 61.5\% & 67.8\% \\
    & \cite{moosavi2018using} & 71.2\% & 79.4\% & 75.0\% & 59.3\% & 69.7\% & 64.1\% & 56.5\% & 64.0\% & 60.0\% & 66.4\% \\
    & \cite{meng-rumshisky-2018-triad} & 84.9\% & 77.3\% & 80.9\% & 60.4\% & 71.8\% & 65.7\% & 44.4\% & 59.2\% & 50.8\% & 65.8\% \\
    \hline
\multirow{7}{*}{\rotatebox[origin=c]{90}{\parbox[c]{2cm}{\centering Latent Structure}}} 
    & \cite{khosla2020using}\footnote{\label{footnote:thank you} We would like to thank the authors for providing the detailed results of their approach.} & \underline{\textbf{92.8}}\% & \underline{\textbf{92.3}}\% & \underline{\textbf{92.2}}\% & \underline{\textbf{82.5}}\% & \underline{\textbf{86.4}}\% & \underline{\textbf{84.4}}\% & 78.9\% & \underline{\textbf{80.7}}\% & \underline{\textbf{79.9}}\% & \underline{\textbf{85.8}}\% \\
    & \cite{wu-etal-2020-corefqa} & 88.6\% & 87.4\% & 88.0\% & 82.4\% & 82.0\% & 82.2\% & \underline{\textbf{79.9}}\% & 78.3\% & 79.1\% & 83.1\% \\
    & \cite{xu2020revealing} & 85.9\% & 85.5\% & 85.7\% & 79.0\% & 78.9\% & 79.0\% & 76.7\% & 75.2\% & 75.9\% & 80.2\% \\
    & \cite{liu2020improving} & 84.5\% & 83.1\% & 83.8\% & 76.2\% & 74.1\% & 75.1\% & 74.0\% & 70.5\% & 72.2\% & 77.0\% \\
    & \cite{kantor2019coreference} & 82.6\% & 84.1\% & 83.4\% & 73.3\% & 76.1\% & 74.7\% & 72.4\% & 71.1\% & 71.8\% & 76.6\% \\
    & \cite{subramanian2019improving}\footnoteref{footnote:thank you}  & 82.4\% & 79.1\% & 80.7\% & 73.7\% & 68.6\% & 71.1\% & 69.7\% & 66.2\% & 67.9\% & 73.2\% \\
    & \cite{aralikatte-etal-2019-rewarding}\footnoteref{footnote:thank you} & 83.0\% & 78.4\% & 80.6\% & 74.5\% & 67.8\% & 71.0\% & 69.4\% & 66.0\% & 67.7\% & 73.1\% \\
    & \cite{lee2018higher} & 81.4\% & 79.5\% & 80.4\% & 72.2\% & 69.5\% & 70.8\% & 68.2\% & 67.1\% & 67.6\% & 73.0\% \\
    \hline
\multirow{5}{*}{\rotatebox[origin=c]{90}{\parbox[c]{1cm}{\centering Entity Based}}} 
    & \cite{xia2020revisiting} & \underline{85.7}\% & 84.9\% & \underline{85.3}\% & \underline{78.1}\% & 77.5\% & \underline{77.8}\% & \underline{76.2}\% & 74.2\% & \underline{75.2}\% & \underline{79.4}\% \\ 
    & \cite{toshniwal2020learning}\footnoteref{footnote:thank you} & 83.2\% & \underline{86.2}\% & 84.7\% & 74.9\% & \underline{78.9}\% & 76.8\% & 70.0\% & \underline{76.7}\% & 73.2\% & 78.2\% \\
    & \cite{clark-manning-2016-deep} & 79.2\% & 70.4\% & 74.6\% & 69.9\% & 58.0\% & 63.4\% & 63.5\% & 55.5\% & 59.2\% & 65.7\% \\
    & \cite{clark-manning-2016-improving} & 78.9\% & 69.8\% & 74.0\% & 70.0\% & 57.0\% & 62.9\% & 62.5\% & {55.8}\% & 59.0\% & 65.3\% \\
    & \cite{wiseman2016learning} & 77.5\% & 69.6\% & 73.4\% & 66.8\% & 57.0\% & 61.5\% & 62.1\% & 53.9\% & 57.7\% & 64.2\% \\ 
    \hline
\multirow{4}{*}{\rotatebox[origin=c]{90}{\parbox[c]{1.7cm}{\centering Mention Ranking}}} 
    & \cite{zhang-etal-2018-neural-coreference-biaffine} & \underline{82.1}\% & \underline{73.6}\% & \underline{77.6}\% & \underline{73.1}\% & \underline{62.0}\% & \underline{67.1}\% & \underline{67.5}\% & 53.0\% & \underline{62.9}\% & \underline{69.2}\% \\
    & \cite{lee2017end} & 81.2\% & {73.6}\% & 77.2\% & 72.3\% & 61.7\% & 66.6\% & 65.2\% & \underline{60.2}\% & 62.6\% & 68.8\% \\
    & \cite{gu2018study} & 79.3\% & 73.9\% & 76.5\% & 70.2\% & {62.7}\% & 66.2\% & 63.5\% & {61.2}\% & 62.3\% & 68.4\% \\
    & \cite{wiseman2015learning} & 76.2\% & 69.3\% & 72.0\% & 66.0\% & 55.8\% & 60.5\% & 59.4\% & 54.9\% & 57.1\% & 63.4\% \\
    \hline
    \hline
\multirow{2}{*}{\rotatebox[origin=c]{90}{\parbox[c]{1.5cm}{\centering Unre-\\stricted}}} 
    & \cite{plu2018sanaphor++} & 65.8\% & 74.7\% & 70.0\% & 58.8\% & 62.4\% & 60.6\% & 52.5\% & 58.6\% & 55.4\% & 62.0\% \\
    & \cite{clark-manning-2016-deep}\footnote{Results from Stanford CoreNLP Framework, which is based on the approach described in \cite{clark-manning-2016-deep} without the limitations that exist in the CoNLL dataset.} & 64.3\% & 72.9\% & 68.3\% & 57.4\% & 60.9\% & 59.1\% & 52.1\% & 58.2\% & 55.0\% & 60.8\% \\
\end{tabular}}
\label{tab:neural_results}
\end{table}

The results presented in Table \ref{tab:neural_results} are only referring to the best results of each implementation previously described, under the best set of parameters, including ensembles. We consider best results the ones that have achieved the highest CoNLL score, regardless of cases where experiments show higher Precision or Recall in one of the individual scores of any of the metrics. As the results are organised in a per model-type basis, the bold results are the best overall in a per-score basis and underline the best results per model-type. As the Mention-Pair models pre-date the publication of the CoNLL 2012 dataset and have been evaluated on different datasets, we do not report a performance comparative to the other models.  

{From the results it is apparent that the best overall results are from latent entity approaches, with the three best performing approaches all deploying different methodologies. }\cite{xu2020revealing}{ is using a merging cluster approach, }\cite{wu-etal-2020-corefqa}{ completely reformulates the task as a QA task and }\cite{khosla2020using}{ is using entity type information. It is important to note that the results reported for }\cite{khosla2020using}{ are acquired using the gold entity type present in CoNLL 2012, with a less than 1\% difference from the scores when having a BERT based model predict the entity types. As a result, we consider both }\cite{khosla2020using}{ and }\cite{wu-etal-2020-corefqa}{ as the current state-of-the-art systems. Furthermore, all of these systems are based on }\cite{joshi-etal-2019-bert}{ which in turn is an extension of }\cite{lee2017end}{.}  

{It is noteworthy that the majority of the proposed models are building on top of previous baseline models }\citep{lee2017end,joshi-etal-2019-bert}{ and hence the results are directly related,} while with cases where the change in scores comes from a parameter fine-tuning, e.g. in \cite{clark-manning-2016-deep}, the fine-tuning is done with the use of reinforcement learning methods. Furthermore, all of the approaches, with the exception of \cite{wiseman2015learning}, \cite{wiseman2016learning} and the early Mention-Pair models, are using dropout to avoid overfitting, and one or more type of word embeddings. Specifically, in \cite{clark-manning-2016-improving,clark-manning-2016-deep} they are pretraining word2vec embeddings \citep{mikolov2013distributed} on the Gigaword corpus and Polyglot embeddings \citep{al-rfou-etal-2013-polyglot}.{ In }\cite{lee2017end}{ and all implementations that are based on this prior to 2018, are using a combination of Glove }\citep{pennington2014glove}{ and CNN character embeddings which are extended by the use of ELMo embeddings }\citep{Peters:2018}{ in }\cite{lee2018higher}{. Later, they are also extended to use BERT }\citep{joshi-etal-2019-bert}{ or SpanBERT }\citep{joshi2019spanbert}{ depending on what the systems used as a base methodology. Similarly, }\cite{clark-manning-2016-improving,clark-manning-2016-deep}{ used RMSprop }\citep{tieleman2012lecture}{ for parameter optimization during learning, while all other approaches in the majority of implementations use Adam }\citep{kingma2014adam}. 

It is important to note that while FFNN remains the same in all of the implementations, in \cite{lee2018higher} the authors use Highway BiLSTMs instead of simple BiLSTMs. 

{Furthermore, }\cite{clark-manning-2016-deep}{ appears twice in Table }\ref{tab:neural_results}{ along with }\cite{plu2018sanaphor++}{ under the Special Cases model type. This is due to the changes in the evaluation scenario for both these models, under which they did not use dataset specific features such as speaker information. This contributed to the notable drop of 5\% in performance in the case of }\cite{clark-manning-2016-deep}.

\subsection{Pronoun Resolution}
Recently, the subtask of Pronoun Resolution has been trending due to the effects of the Kaggle competition\footnote{https://www.kaggle.com/c/gendered-pronoun-resolution/} and gender bias in both the CoNLL 2012 dataset and sentence representation methodologies. As a result, due to the recency of the subject, no clear benchmark exists and not all approaches to Pronoun Resolution are evaluated in the same dataset. We present the results in Table \ref{tbl:pronoun_conll} in terms of F1 score on CoNLL 2012 datasets for the approaches that were evaluated for Pronoun Resolution and the results for the described implementations of the competition on the GAP dataset for Gender Pronoun Resolution in Table \ref{tbl:pronoun_gap}.

\begin{table}
\centering
\caption{Pronoun Resolution on CoNLL 2012}
 \begin{tabular}{l|l}
 & F1-score  \\
 \hline
 \cite{tenney2019you} &  91.4\% \\
 \cite{zhang2019incorporating} & 81.0\% \\
 \cite{zhang2019knowledge} & 75.5\% \\
 \cite{clark2019does} & 65.0\% \\
\end{tabular}
\label{tbl:pronoun_conll}
\end{table}
\begin{table}
\centering
\caption{Gender Pronoun Resolution on GAP }
 \begin{tabular}{l|llll|l}
& M & F & B & O & logloss\\
\hline 
\cite{attree2019gendered} & 94.0\% & 91.1\% & 0.97\% & 92.5\% & .317\\
\cite{ionita2019resolving} & 92.7\% & 90.0\% & 0.97\% & 91.4\% & .346\\
\cite{liu2019anonymized}\footnote{\label{footnote:thank_you_gender}We would like to thank the author for providing gender scores  after direct contact.} & 91.6\% & 90.8\% & 0.99\% & 89.26\% & .179\\
\cite{xu2019look}\footnoteref{footnote:thank_you_gender} &  79.9\% & 81.1\% & 1.01\% & 80.3\% & .493\\ 
\cite{GapDataset} & 72.8\% & 71.4\% & 0.98\% & 72.1\% & - \\
\cite{lee2017end} & 67.7\% & 60.0\% & 0.89\% & 64.0\% & - \\
\end{tabular}
\label{tbl:pronoun_gap}
\end{table}

We notice that \cite{tenney2019you} manages - using a simple model - to get very high scores, while the use of external knowledge has improved the baseline in \cite{zhang2019incorporating}. While the approach by \cite{clark2019does} has the worst performance, it is important to note that it is coming out of a sentence representation model with attention probing and is not actually a model designed for the task of pronoun resolution, or even fine-tuned for it. 

\section{Discussion}
\label{sec:discussion}
{ Coreference Resolution has made significant leaps in performance in the recent years, boosted by the use of deep neural networks, word embeddings and language modelling. At the first stages of the NN approaches, the focus was to build good foundations for the task, that is, developing neural pairwise scoring function, neural modelling of the problem and establishing a training objective. That enabled later implementation of Entity-Based approaches, following a similar timeline as non-neural approaches. At the same time, the task shifted from using words in their surface forms and feature extraction to using embeddings and end-to-end neural approaches. What significantly boosted performance and has seen the greatest advancements in CR is the use of novel language models joint by latent entity representations.}

In the early approaches, the use of word surface forms and feature extraction has been very beneficial, but a problem in generalization was present. All implementations that adopted such approaches have found that pairwise features (distance and head matching in particular) have contributed the most to the performance boost. Similarly, techniques for mapping linguistic features, such as the one presented in \cite{moosavi2018using} have also greatly helped such approaches, even though they are limited in scope and require a lot of human effort. What is more, these techniques were hindered by the cascading errors introduced by mention detection tools that were required to extract all the mentions and candidate antecedents. With the use of word embeddings and span representation, these approaches were translated into neural functions to learn to identify mentions and pair them together. However, while better generalization was achieved through embeddings, their use increased FP links because they confuse paraphrasing with relatedness. {The addition of ELMo, BERT and SpanBERT embeddings mitigated the issue and provided word representations with semantic context.}

Going past the limits of the underlying methodologies used, the approaches introduced even more limits themselves.

The mention-ranking models that use mention detection tools \citep{wiseman2015learning} improved on identifying non-anaphoric mentions while in models that use spans of text as input \citep{lee2017end,zhang-etal-2018-neural-coreference-biaffine} this is replaced by attention mechanisms, which increases precision scores. However, they fail to make global decisions and, as a result, are prone to errors in transitivity. While the transitivity issue is partially dealt in by a novel clustering methodology during inference which solves incompatible clusters \citep{gu2018study}, these models are unable to make decisions that require world knowledge.

The entity-based models have generally implemented the task as an agglomerative clustering problem, predicting clusters directly. As such they improve on several aspects over the mention-ranking models, although they are also using mention detection tools which makes them prone to cascading errors. The have fewer FL mistakes, which is the main improvement over the previous approaches, leading to better identification of pleonastic pronouns, non-anaphoric pronouns and non-anaphoric nominal mentions. Since all of the approaches use word embeddings, they also achieved a significant improvement over linking nominals with no head match over previous approaches. {The most recent entity-based models focus on constraining the clustering decisions to further increase the performance as well as to save memory.}

The latent structure models are converting the task of Coreference Resolution into a task of predicting latent structures and inferring the clusters from the resulting structures. As a result, the required iterations to create such structures increase complexity, making the process very computationally expensive. {What is more, to increase contextual information, most recent approaches incorporate methods to further enhance the latent space with context in the form of entity equalization }\citep{kantor2019coreference}{, entity type information}\citep{khosla2020using}{ or even completely changing the scope approach to Question Answering }\citep{wu-etal-2020-corefqa}{.} As they attempt to solve the same issues faced by mention-raking problems, in a similar manner as entity-based models, they outperform both model types in terms of precision and recall due to their ability to model very distant connections successfully.

The language modelling approaches attempt to map different language aspects, in order to enhance the CR task and face the issues of the mention ranking models. The approach described in \cite{swayamdipta-etal-2018-syntactic} boosts performance by better defining pronominal mentions, due to the syntactic information, which is also improved by the cross-sentence dependencies that are built in word representations in \cite{luo2018learning-crosssentence}. \cite{meng-rumshisky-2018-triad} improves on the problem of transitivity and salience by explicitly predicting triads instead of dyads, while \cite{moosavi2018using} uses linguistic features to better improve generalization of the models. {However, the improvement that the BERT model adaptation of }\cite{lee2017end}{ described in }\cite{joshi-etal-2019-bert}{, as well as the BERT's objective functions change that was presented in }\cite{joshi2019spanbert}{ to learn span representations, have outperformed all other language modelling approaches. By learning span boundaries and having an extended learning scope over the base model, it effectively understands entities within the spans, while using contextual information to do so. These two approaches have appropriately been used by the majority of following implementations.}

Due to the results of the Kaggle competition in Gendered Pronoun Resolution, advancements have been achieved in removing bias via various techniques. Although the results seem promising, most of the approaches are not reflecting realistic improvement to the task of Entity Coreference Resolution as resolver predictions from the general task were used in the training process. The work on Pronoun resolution on the other hand provides insightful results towards the better use of sentence representation models to the task and the use of external knowledge features. 

Moreover, \cite{moosavi2018using} developed a novel algorithm to automatically extract the minimum span in a variety of datasets to solve this issue and boost performance of the current CR resolvers. As the majority of the implementations discussed use spans, this contribution is very important and can boost their performance. 

\section{Applications of Coreference Resolution}
\label{sec:applications}
{Coreference Resolution has been of increasing research interest due to it's inherited ties to discourse and natural language understanding. What is more the importance of the improvements in the task of Coreference Resolution can be found in its uses in state-of-the-art approaches in a plethora of Natural Language Processing tasks. 

We have already seen implementations in Entity Linking }\citep{kundu2018neural,ling2015design}{, Machine Translation }\citep{popescu2019context,urbizu2019deep,voita2018context}{, Chat bots }\citep{zhu2018lingke,jonell2018fantom}{, Summarization }\citep{song2019abstractive,barros2019natsum}{, Named Entity Recognition }\citep{dai2019coreference,he2020entity}{, Question Answering }\citep{bhattacharjee2020investigating,chen2019multi}{ and Sentiment Analysis }\citep{7975227,le2016sentiment}{ which infuse their models with anaphoric information to achieve better results.}

{The effectiveness and importance of Coreference Resolution and Anaphora Resolution in Sentiment Analysis is explored in depth in }\cite{DBLP:journals/corr/abs-1805-11824}{, which summarizes the different approaches in which the anaphoric information have been used to enhance the task. }\cite{lata2020comprehensive}{ and }\cite{saunders2020neural}{ describe the need for anaphoric information in summarization and its impact. The common ground between the two studies is that anaphoric information plays a key role in the better representation of the source information, which in turns has a positive effect in the end tasks.}

{With the recent developments in CR, due to the cornucopia of approaches introduced in the last few years, more methodologies can incorporate anaphoric information through different architectures or even make explicit entity decisions. }\cite{luan-etal-2018-multi}{ utilizes CR to enhance Knowledge Graphs, which in turn can be used for graph based systems }\citep{wang2020word,correa2018word,de2016using,krishna2016extractive}{. Furthermore, Language Modelling approaches with explicit entity decisions have been of research interest }\citep{stylianou-vlahavas-2020-e,kunz2019entity,ji2017dynamic,yang2016reference}{, which attempt primitive cluster ranking to define what entity the predicted word is part of. These type of models can greatly benefit from novel CR systems to increase their accuracy and overall performance. What is more, they offer an easy means to incorporate entity information within the word representations that can be adopted by all Natural Language Processing tasks that use such a model as the source of their respective word embeddings. }

\section{Challenges and future scope}
\label{sec:challenges}
{Coreference Resolution has made significant strides through the various neural approaches. On the CoNLL 2012 dataset alone, approaches have seen a 22\% increase in Average F1 the past 5 years. However, it is evident that CR has been a task of significant difficulty, with challenges remaining to be dealt with in both research and in practice.}

{Generalization is a big issue in the task of CR, and the performance of the systems is not reflecting the reality. The flaws of the CoNLL dataset, which is used as the benchmark for the task, along with the bias in the scoring metrics (discussed in Section }\ref{sec:metrics}{) is hindering improvement. In section }\ref{sec:datasets}{ we highlighted the issue with lexical features in the CoNLL dataset, where all the systems that use surface form of mentions instead of spans are affected by it. What is more, the CoNLL dataset does not completely adhere to the definition of CR, as described in section }\ref{sec:differences}{. Furthermore, the gender bias is also a very important issue }\citep{rudinger-etal-2018-gender,chen2013linguistically}{. While dealing with gender disambiguation, the performance of }\cite{lee2018higher}{ decreases rapidly in terms of recall across all scores as reported in }\cite{subramanian2019improving}{. It was also discovered that not only Coreference Resolution data is biased but also the representations that derive from the use of ELMo and BERT }\citep{kurita2019measuring,zhao2019gender,clark2019does}{. }\cite{cao-daume-iii-2020-toward}{ offers a detailed analysis of the effects of gender bias and provides a technique of measuring gender bias in coreference resolution annotations and Natural Language processing tasks in general. }

{Due to the results of the Kaggle competition in Gendered Pronoun Resolution, advancements have been achieved in removing bias via various techniques. Although the results seem promising, most of the approaches are not reflecting realistic improvement to the task of Entity Coreference Resolution as resolver predictions from the general task were used in the training process. The work on Pronoun resolution on the other hand provides insightful results towards the better use of sentence representation models to the task and the use of external knowledge features.}

{Most noticeably the majority of the approaches, while novel in various aspects, where published around the same time period. As a result, the improvements that have been introduced are not adopted to the best possible baseline. The majority of the language modelling methods are infused using }\cite{lee2017end}{ as their baseline, which is outperformed by }\cite{joshi-etal-2019-bert}{ with the change of ELMo to BERT embeddings and as well as present two approach for segmentation handling. Although, some of the methodologies do adopt similar methods, it appears that this was done out of individual progression without mentions to the work done in }\cite{joshi-etal-2019-bert}{. Moreover, }\cite{moosavi2018using}{ developed a novel algorithm to automatically extract the minimum span in a variety of datasets to solve this issue and boost performance of the current CR resolvers. As a significant amount of implementations discussed use spans, this contribution is very important and can boost their performance. Similarly, approaches that used BERT and were based on }\cite{lee2018higher}{, were proven to have suboptimal performance in }\cite{xu2020revealing}{ due to second order inference. What is more, with the exception of }\cite{wu-etal-2020-corefqa}{ and }\cite{liu2020improving}{, all of the proposed methodologies only consider candidate antecedents once and are unable to model cataphoric phenomena by only doing forward passes. Detailed reviews such as this one, provide a thorough description of the landscape, preventing researchers from reiterating already established methodologies.}

{Coreference Resolution and its subtasks are also being set back by the lack of an agreed-upon standard for both the datasets and evaluation metrics }\citep{poesio2016anaphora}{. The CoNLL 2012 dataset has clear flaws, and even though it has served as the benchmark to define the state-of-the-art, the issues hinder improvement. Specifically, singletons are not explicitly labelled, there is a big overlap between the standarized test, development and train splits and there is also the issue of gender bias. To make matters worse, the cases where the resolution of pronouns is not based on surface cues like number and gender are scarce, leading to a flawed sense of state-of-the-art performance without any real contributions in the ``uphill battle'' of using world knowledge for coreference predictions. The recently proposed LitBank dataset can serve as an cross-domain benchmark, and we have noticed some of the most recent works to use it as such, hinting to the need for further ways of measuring performance. Even so, it is not a suitable replacement for the CoNLL 2012.}

{The use and reliability of the performance metrics used in Entity Coreference Resolution is also questioned }\citep{LEA:P16-1060,agarwal-etal-2019-evaluation}{ as there are significant flaws in all the currently used ones, that are not solved by their parallel usage. We strongly propose the combination of LEA, NEC and CEAF metrics for the task of unrestricted coreference resolution as well as Consistency for Pronoun Resolution. We believe that LEA is a strong replacement for the MUC score, while NEC highlights the importance of resolving Named Entities, something that has been overlooked by other metrics. A similar to the CoNLL unweighted average F1 score can also be calculated from the respective F1 measures. Consistency is also important in the task of Pronoun Resolution to allow for a deeper evaluation of the decisions made by the resolvers.}

{Going past the academic challenges and flaws in the approaches, practical obstacles also need to be tackled to increase the real world applications of Coreference Resolution, and hence unrestricted Coreference Resolution. 
Currently, CR is hindered by the available resources. The CoNLL 2012 corpus can no longer be used as the benchmark for CR resolvers in its own. However, creating a dataset that can support deep neural architecture training and is of high quality is difficult and costly. Using a unifying coreference annotation scheme and methodology during the making of such resources can be crucial to the final quality. As such, the annotation scheme proposed in }\cite{prange2019semantically}{ and the annotation methodology }\citep{aralikatte2019model}{ are recommended, to ensure consistency. In order to decrease the cost, a combination of the crowd sourcing suite specifically for CR annotations developed by }\cite{bornstein2020corefi}{ and the proposed active learning approach to only require annotation of the hard coreference decisions in }\cite{li-etal-2020-active}{ can contribute significantly.}

{Furthermore, the current CR systems are very demanding in terms of both physical resources (memory) and computation resources (computational complexity), making them unsuitable for incorporation with other tasks in a joint learning approach. Novel methodologies have started to consider such limitations }\citep{xia2020revisiting,toshniwal2020learning}{, attempting to constrain the memory requirements and the computations required to build coreference clusters. We believe that work in enhancing the training objectives such as the one presented in }\cite{le-titov-2017-optimizing}{ will hold key roles to such advancements.}

\section{Conclusions} 
\label{sec:conclusions}
{Coreference Resolution is a very important part of discourse and by extension of language modelling and language understanding. Although it has seen great progress through the use of neural networks, it is far from solved and it is considered as one of the most difficult tasks due to the required world knowledge and inference problems that surround it. The aim of this review is to provide a detailed review of the task and the methodologies used, identify the weaknesses and enable better and targeted future research that would allow for its faster progression.}

{In this survey we reviewed the different neural approaches and categorized them based on their respective approach to the Coreference Resolution task. As a result, the models types are identified as either Mention-Pair, Mention-Ranking, Entity-Based, Latent-Structure or Language-Modelling for which their merits and inferiorities are outlined. Moreover, we identified the issues with the current evaluation metrics and resources used and proposed alternatives that can be directly applied. We further defined the methodologies that should be followed for the creation of better resources, which are clearly needed for this task, as well as approaches to do so more efficiently. Consequently, we take a step towards solidifying agreed-upon standards for both metrics and evaluations that the task is missing.}

{Additionally, this survey also briefly discussed the advancements in Pronoun Resolution, an important subtask of Corefrence Resolution. We identified its advancements and how they can be applied to Coreference Resolution, while also focusing on the issue of Gender Pronoun Resolution and the overall gender bias that exists in both systems and resources.}

{We conclude this survey with an thorough overview of the applications of Coreference Resolution that are either currently being researched or that have been enabled through recent developments and the current challenges from an academic and practical scope. Finally, we list the future steps required to enhance Coreference Resolution as a whole.}


\section*{Acknowledgements}
This research is co-financed by Greece and the European Union (European Social Fund- ESF) through the Operational Programme ``Human Resources Development, Education and Lifelong Learning'' in the context of the project “Strengthening Human Resources Research Potential via Doctorate Research” (MIS-5000432), implemented by the State Scholarships Foundation (ΙΚΥ).

\bibliography{mybibfile}

\end{document}